\documentclass[10pt,twocolumn,letterpaper]{article}

\usepackage{cvpr}              
\usepackage{multirow} 
\usepackage{soul} 
\usepackage{amssymb}
\usepackage{pifont}
\usepackage{bbm}
\usepackage{mathtools}
\usepackage{amsmath}
\usepackage{makecell}

\usepackage{xcolor}
\usepackage{soul}
\usepackage{makecell}
\usepackage{dsfont}
\usepackage{bm}
\usepackage{subfloat}
\usepackage{color, colortbl}
\usepackage{soul}

\newcolumntype{a}{>{\columncolor{yellow}}c}
\newcolumntype{b}{>{\columncolor{white}}c}

\newcommand{\shortname}{PnP-OVSS}
\usepackage[accsupp]{axessibility} 

\definecolor{cvprblue}{rgb}{0.21,0.49,0.74}
\usepackage[pagebackref,breaklinks,colorlinks,citecolor=cvprblue]{hyperref}

\def\paperID{12308} 
\def\confName{CVPR}
\def\confYear{2024}

\title{Emergent Open-Vocabulary Semantic Segmentation from Off-the-shelf Vision-Language Models}

\author{Jiayun Luo$^\flat$, Siddhesh Khandelwal$^\sharp$, Leonid Sigal$^\sharp$, and Boyang Li$^\flat$\\
$^\flat$Nanyang Technological University, Singapore\\
$^\sharp$University of British Columbia, Vector Institute for AI, Canada\\
{\tt\small \{luoj0028, boyang.li\}@ntu.edu.sg, \{skhandel, lsigal\}@cs.ubc.ca}
}

\begin{document}

\twocolumn[{%
\renewcommand\twocolumn[1][]{#1}%
\maketitle
\begin{center}
    \centering
    \captionsetup{type=figure}
    \includegraphics[width=1\textwidth]{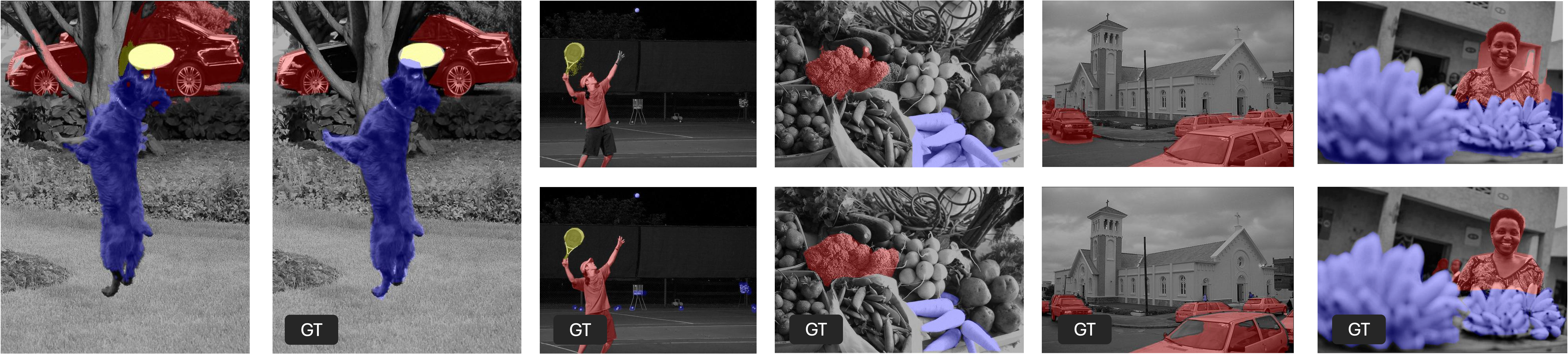}
    \vspace{-0.25in}
    \captionof{figure}{{\bf Qualitative Results of PnP-OVSS + BLIP.} Images are from Pascal VOC and COCO Object. The right columns and bottom rows show the ground-truth (GT); the rest are our results. Note the good results on small objects like the frisbee and the tennis racket.}
\end{center}
}]



\begin{abstract}
\vspace{-0.2in}
From image-text pairs, large-scale vision-language models (VLMs) learn to implicitly associate image regions with words, which prove effective for tasks like visual question answering. 
However, leveraging the learned association for open-vocabulary semantic segmentation remains a challenge.
In this paper, we propose a simple, yet extremely effective, training-free technique, Plug-and-Play Open-Vocabulary Semantic Segmentation (PnP-OVSS) for this task. PnP-OVSS leverages a VLM with direct text-to-image cross-attention and an image-text matching loss. To balance between over-segmentation and under-segmentation, we introduce Salience Dropout; by iteratively dropping patches that the model is most attentive to, we are able to better resolve the entire extent of the segmentation mask. \shortname{} does not require any neural network training and performs hyperparameter tuning without the need for any segmentation annotations, even for a validation set. PnP-OVSS demonstrates substantial improvements over comparable baselines (+26.2\% mIoU on Pascal VOC, +20.5\% mIoU on MS COCO, +3.1\% mIoU on COCO Stuff and  +3.0\% mIoU on ADE20K). Our codebase is at \url{https://github.com/letitiabanana/PnP-OVSS}.


\end{abstract}    
\section{Introduction}
\label{sec:intro}

The classic task of 
semantic segmentation \cite{guo2018review, hao2020brief} aims to classify pixels to their object types. Traditional supervised methods require dense pixel-level annotations and are restricted to recognizing a predefined set of objects. To relax these constraints, open-vocabulary semantic segmentation \cite{ghiasi2022scaling, xu2023side, xu2023learning, liang2023open, xu2022simple, luo2023segclip,qin2023freeseg,mukhoti2023open, zhang2023simple,han2023global,chen2023exploring,zabari2022open} aspires to identify arbitrary object categories, whereas weakly supervised techniques \cite{zhou2022regional,li2022expansion,chen2022class,lee2022weakly,jiang2022l2g,chen2022self,ru2023token,xu2022multi,du2022weakly,ru2022learning,wang2022looking} can acquire pixel-level localization capabilities from coarse supervision, {\em e.g.}, image labels or boxes. 

Large-scale vision-language models (VLMs) pretrained on image-text pairs \cite{Singh_FLAVA_2022,Zhang_VinVL_2021,OSCAR-2020,li2022blip,li2023blip2,xu2022bridgetower,flamingo2022} achieve unprecedented performance on multimodal tasks, such as describing arbitrary images and answering free-form, open-ended questions about them (either with \cite{Singh_FLAVA_2022,li2022blip,li2023blip2} or without finetuning \cite{flamingo2022,Tiong-PnPVQA-2022,guo2023images,xenos2023simple}). These tasks apparently involve some ability to localize objects. For example, to answer the question ``{\em what objects appear on the table?}'', the model would have to first localize the table in the image and identify the objects on it. Hence, it is reasonable to conjecture that the VLM network learns to perform open-vocabulary localization from image-text pretraining. However, distilling the localization capability from the VLMs remains an open challenge.

Most existing methods for open-vocabulary semantic segmentation (OVSS) from VLMs usually obtain single vector encodings for the visual and text inputs respectively \cite{xu2023learning,liu2022open,ranasinghe2023perceptual,xu2022groupvit,luo2023segclip,ren2023viewco}. However, pooling every token into a single vector likely discards information about detailed positions of objects and words. We investigate the use of pretrained cross-attention layers for OVSS, which retain finer-grained correspondence between text and image patches. 

\begin{figure}
\captionsetup[subfloat]{justification=centerfirst,indention=10pt,format=hang,singlelinecheck=false}
\centering
\subfloat{\includegraphics[width=.245\linewidth,height=.245\linewidth]{Figures/ElephantOg.jpeg}}
\hfill
\subfloat{\includegraphics[width=.245\linewidth,height=.245\linewidth]{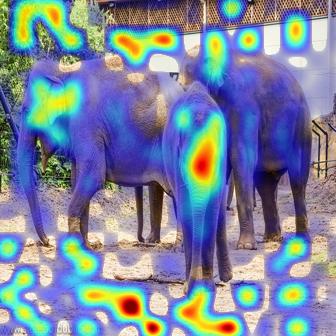}}
\hfill
\subfloat{\includegraphics[width=.245\linewidth,height=.245\linewidth]{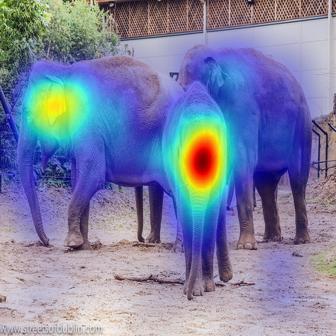}}
\hfill
\subfloat{\includegraphics[width=.245\linewidth,height=.245\linewidth]{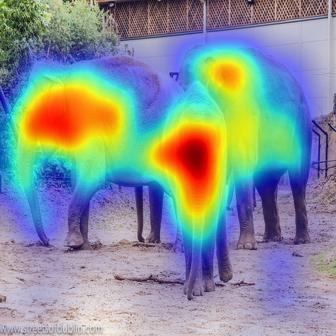}}

\setcounter{subfigure}{0}
\subfloat[Original\\image]{\includegraphics[width=.245\linewidth,height=.245\linewidth]{Figures/Revised_Giraffe_ORG.jpeg}}
\hfill
\subfloat[Cross- \protect\\ \, attention]{\includegraphics[width=.245\linewidth,height=.245\linewidth]{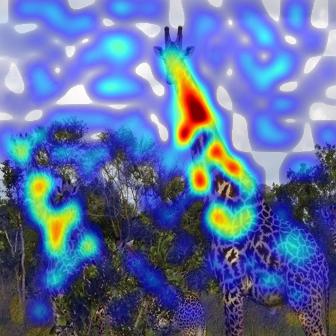}}
\hfill
\subfloat[GradCAM]{\includegraphics[width=.245\linewidth,height=.245\linewidth]{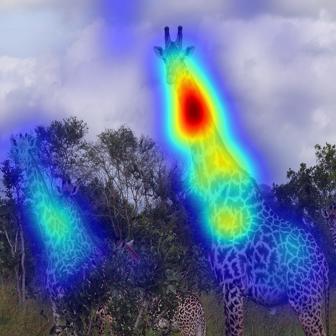}}
\hfill
\subfloat[Salience \\ DropOut]{\includegraphics[width=.245\linewidth,height=.245\linewidth]{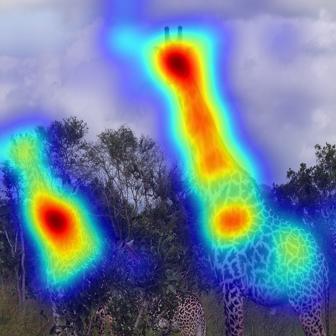}}

\caption{Segmentation masks for \emph{elephant} and \emph{giraffe} using (a)~off-the-shelf cross-attention, (b) cross-attention + GradCAM, and (c) cross-attention + GradCam + Salience DropOut (\S \ref{sec:salience-dropout}). The naive cross-attention masks are too inclusive whereas GradCAM is too exclusive. }\label{fig:attention}
\vspace{-0.15in}
\end{figure}

Nevertheless, a naive application of cross-attention between the class name and the image patches lead to overly broad segmentation masks that include irrelevant parts of the image ({\em i.e.}, over-segmentation, see Fig. \ref{fig:attention} (b)). To alleviate this, \cite{Tiong-PnPVQA-2022} employs gradient information from the image-text matching loss, 
GradCAM\cite{GradCAM2017}-style, to sharpen the attention-based masks for the purpose of guiding caption generation. However, this results in masks that capture only the most discriminative regions of an object, such as the head of an elephant ({\em i.e.}, under-segmentation, Fig. \ref{fig:attention} (c)). To acquire complete object masks, we propose Salience DropOut, which iteratively drops the image patches with high GradCAM attention scores, forcing the model to attend to less discriminative but relevant object parts.

Another important consideration is the cross-attention layer and the attention head to extract the masks from. These hyperparameters have enormous influence on the final results and are traditionally tuned on a validation set with pixel-level mask annotations. 
To eliminate the need for dense annotations, we propose an weakly-supervised reward function based on CLIP \cite{radford-clip-2021}. On a validation set with images as well as object class names, the technique contrasts the extracted object regions with a blank image. If, according to CLIP, the former is more similar to the corresponding class name than the latter, we increment the reward. 
All hyperparameter tuning of our technique is performed with a simple random search with this reward, leading to high performance. 

\fussy{
In summary, we propose Plug-and-Play Open-vocabulary Semantic Segmentation (PnP-OVSS), an extremely simple and training-free framework to extract semantic segmentations from VLMs. At zero extra training cost, PnP-OVSS can be used with
any pretrained VLM with text-to-image cross attention layer and an image-text matching loss. It has zero reliance on pixel-level annotations, including a validation set for hyperparameter tuning. At the same time, PnP-OVSS delivers excellent performance. It not only beats the training-free baseline with remarkable margins (+26.2\% mIoU on Pascal VOC, +20.5\% mIoU on MS COCO, +3.1\% mIoU on COCO Stuff and  +3.0\% mIoU on ADE20K), but also outperforms most recent techniques within the past two years that require extensive finetuning on top of the VLM pretraining.}

With this paper, we make three contributions:
\begin{itemize}
  \item We propose to combine text-to-image attention, GradCAM, and Salience DropOut to iteratively acquire accurate
  segmentation of arbitrary classes from pretrained VLM.
  \item We replace the densely annotated validation set for hyperparameter tuning, which is needed by most existing methods, with a contrastive reward function based on CLIP. This reward function, coupled with random search, finds a good set of hyperparameters for OVSS. 
  \item The proposed method, PnP-OVSS, is simple to use, requires no extra finetuning, and delivers high performance. Its success hints at a new direction for open-vocabulary segmentation tasks leveraging large VLMs. 
\end{itemize}



\section{Related Work}
\label{sec:related}

\subsection{Large-Scale Vision-Language Model}

Large-scale vision-language models (VLMs), trained on millions of image-text pairs, have become the foundation for many multimodal tasks. Architecturally, the straightforward approach to training such methods involves aligning visual and textual latent representations via a simple dot product \cite{radford-clip-2021, Jia-ALIGN-2021}. However, this is insufficient for complex structured tasks like visual question answering or image captioning, which require specialized approaches that employ separate encoders before cross-attention between modalities \cite{ALBEF,li2022blip,xu2022bridgetower,li-etal-2022-mplug,yu2022coca} or self-attention networks over all tokens from both modalities \cite{Wonjae-ViLT-2021,OSCAR-2020,Singh_FLAVA_2022,li-etal-2021-unimo,UNITER-2020}. Another design dimension when training VLMs is the loss function. Commonly used losses include image-text contrastive learning \cite{radford2021learning,ALBEF,li2022blip,li2022mplug,zeng2021multi, Singh_FLAVA_2022}, image-text matching (ITM) \cite{ALBEF,li2022blip, xu2022bridgetower,li2022mplug,zeng2021multi,Singh_FLAVA_2022,Wonjae-ViLT-2021,xu2023mplug,zeng2022x}, prediction of masked tokens or patches \cite{xu2022bridgetower,li2022mplug,zeng2021multi,Singh_FLAVA_2022}, and language modeling \cite{chen2022visualgpt,li2022mplug,flamingo2022,li2022blip}.

This work utilizes models with unimodal encoders followed by cross-attention fusion \cite{li2022blip,xu2022bridgetower,zeng2021multi,li2022mplug,ALBEF,xu2023mplug,zeng2022x}, as they work with high-level features, and accurately attend to the appropriate image patches. Additionally, we utilize the gradient from the ITM loss in the GradCAM step (\S \ref{sec:gradcam}) to sharpen segmentation masks.


\subsection{Zero-shot Semantic Segmentation}


\begin{table}
  \centering
  \label{main result}
  \small
  \begin{tabular}{@{}llr@{}}
    \toprule
    Method  & PT Data Size & FT data size \\
    
    \midrule
    \multicolumn{3}{c}{\textit{Requires finetuning on image-text pairs}}\\
    OVSegmentor \cite{xu2023learning} & - & 4.3M \\ 
    Vil-Seg \cite{liu2022open}  & 400M & 412M \\
    GroupVit* \cite{xu2022groupvit} & - & 3.4M\\
    GroupVit \cite{xu2022groupvit} & -  & 26M\\
    CLIPpy \cite{ranasinghe2023perceptual}   & - & 134M   \\
    SegClip \cite{luo2023segclip}  & 400M & 3.4M \\
    ViewCo \cite{ren2023viewco} & - & 26M\\
    TCL \cite{cha2023learning} & 400M & 15M\\
    PACL \cite{mukhoti2023open} & 400M  & 30M \\
    \midrule
    \multicolumn{3}{c}{\textit{Requires finetuning but not image-text pairs}}\\
    
    MaskClip w/ ST \cite{zhou2022extract} & 400M  & 1.2M\\
    ZeroSeg* \cite{chen2023exploring}  & 400M  & 3.4M \\
    ZeroSeg \cite{chen2023exploring} & 400M & 1.2M \\
    \midrule
    \multicolumn{3}{c}{\textit{Requires no finetuning}}\\
    MaskClip\cite{zhou2022extract}  & 400M  & 0 \\
    Reco\cite{shin2022reco}  & 400M  & 0 \\
    PnP-OVSS (Ours)   &   &  \\
    \;\; + BLIP & 129M & 0 \\
    \;\; + BridgeTower & 400M+ 4M  & 0 \\
    
    \bottomrule
  \end{tabular}
  \caption{Training data of current Zero-shot semantic segmentation  methods with only text supervision. PT stands for pretraining with image-caption data, FT stands for fine-tuning with image-caption data. ST stands for self training. We list only the image-caption data used for pretraining and all type of data for finetuning. For hyperparameter tuning, our method uses CLIP-L/14 pretrained with 400M data to calculate the reward.}
  \label{tab:related work}
\end{table}

\sloppy{Zero-shot semantic segmentation predicts a dense segmentation mask for any object class described by a given text prompt, with only prior exposure to class-agnostic image-level supervision. This contrasts with weakly supervised semantic segmentation \cite{zhou2014object, bazzani2016self,cinbis2016weakly, ru2023toco,kolesnikov2016seed,tang2018regularized,zhang2020reliability,ke2021universal,wei2017object,hou2018self,wei2017object,kumar2017hide,zhang2018adversarial}, which relies on class-specific annotations, and unsupervised object discovery \cite{simeoni2021localizing,darcet2023vision,wang2022self,wei2019unsupervised,vo2021large,cho2015unsupervised,rubinstein2013unsupervised,vo2019unsupervised}, which identifies the sole object in the foreground.} 

Traditional methods \cite{xian2019semantic,bucher2019zero,gu2020context,pastore2021closer} train a classifier to distinguish between seen and unseen visual features, wherein the unseen visual features are obtained from a generative model, trained on pairs of seen class image-text embeddings. Recently, methods additionally leverage knowledge from VLMs to attain better matching of visual and textual features \cite{ghiasi2022scaling,xu2022simple,li2022languagedriven, zhou2022extract,liang2023open}. Concretely, they train the segmentation network with dense annotations, while replacing a part of the framework with components from VLMs. 

Recently, methods have explored the use of additional supervision to further reduce the need for pixel level annotations. Prior work is compared in Table \ref{tab:related work}.

\vspace{0.05in}
\noindent\textbf{Models Finetuned on Image-Text Pairs.} Methods proposed in \cite{xu2022groupvit,ren2023viewco,luo2023segclip,xu2023learning,liu2022open,ranasinghe2023perceptual,mukhoti2023open,cha2023learning} require paired image-text annotations to adapt to the zero-shot segmentation task. Specifically, approaches in \cite{xu2022groupvit,ren2023viewco,luo2023segclip,xu2023learning,liu2022open,ranasinghe2023perceptual} cluster semantically similar pixels by contrasting grouped region embeddings with textual embeddings, usually via the use of contrastive and self-supervised losses. PACL \cite{mukhoti2023open} modifies the constrastive loss to operate over aggregated patch embeddings (instead of image embeddings) to encourage better alignment between image patches and text. TCL \cite{cha2023learning} achieves a similar patch-text alignment by introducing an additional module to extract text-grounded image regions.

\vspace{0.05in}
\noindent\textbf{Models Finetuned on Pseudo-labels.} Methods in \cite{chen2023exploring, zhou2022extract} do not require additional image-text supervision, but involve additional fine-tuning to adapt to the segmentation task. ZeroSeg \cite{chen2023exploring} attempts to match its own group embeddings with multi-scaled segment embeddings obtained from CLIP \cite{radford-clip-2021}. MaskClip w/ ST \cite{zhou2022extract} modifies the global attention pooling layer within CLIP to output segmentation masks, which are used as pseudo-labels to train a segmentation network.

In comparison, our proposed PnP-OVSS does not require \emph{any fine-tuning or additional paired image-text annotations}. It can directly distill high quality open vocabulary semantic segmentations from any VLM with direct text-to-image cross-attention and an image-text matching loss \cite{li2022blip,xu2022bridgetower,zeng2021multi,li2022mplug,ALBEF,xu2023mplug,zeng2022x}. Compared to approaches in \cite{shin2022reco, zhou2022extract} that perform zero-shot open vocabulary semantic segmentation under a similar no training and no additional annotations paradigm, PnP-OVSS achieves considerably superior performance.

\begin{figure*}[t]
\centering
  \includegraphics[width=1\linewidth]{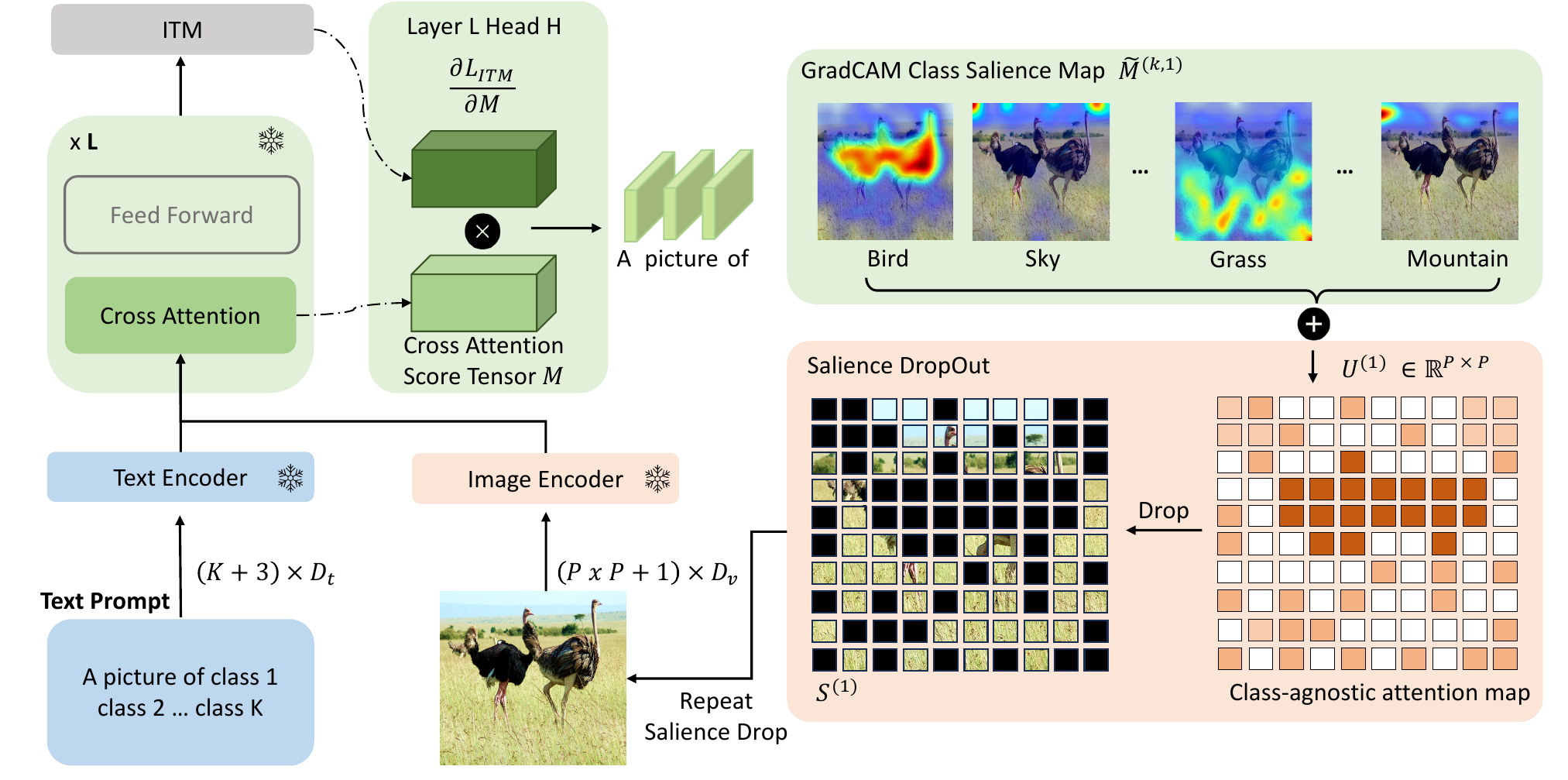}
    \caption{The first iteration with cross-attention + GradCAM + Salience DropOut. The text prompt contains $K$ class names and the image contains $P\times P$ patches. 
    From a cross-attention layer and an attention head in the pretrained VLM, we obtain $K$ attention score maps of size $P\times P$, which are sharpened by GradCAM using gradients from the image-text-matching (ITM) loss. To get more complete predictions, we perform Salience Dropout, which repeatedly zero out image patches of the highest average scores and feeds the remaining patches to the image encoder again, forcing the model to attend to other less discriminative patches. We show example salience maps from all iterations in Fig. \ref{fig:short-b}.}
    \label{fig:short-a}
\end{figure*}

\section{Method}
\label{sec:method}


\begin{figure}
  \centering
    \includegraphics[width=\columnwidth]{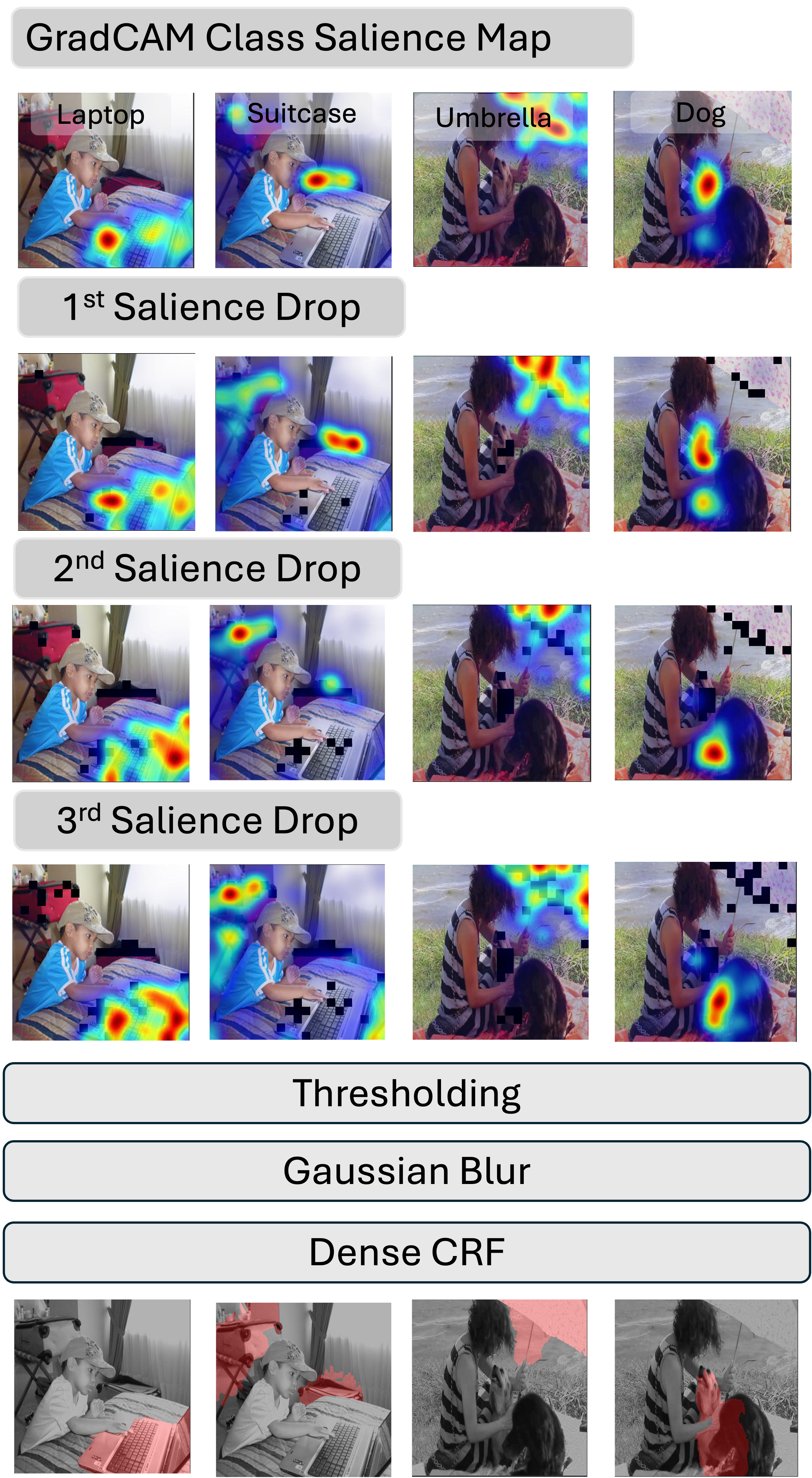}
    \caption{An illustration of Salience DropOut, showing GradCAM salience values after each iteration. Black squares in the images indicate dropped patches. We obtain the final result by summing the salience maps from all iterations and applying thresholding, Gaussian blur, and Dense CRF.   }
    \label{fig:short-b}
\end{figure}

\shortname{} has five major steps. First, we prompt GPT4o to output the categories appear in each image. Second, we extract a cross-attention salience map for each predicted category from a VLM. Third, we sharpen the salience map by weighing it with the ITM gradient in the style of GradCAM. Fourth, we apply Salience DropOut that iteratively completes the salience maps. Finally, we apply Dense CRF \cite{krahenbuhl2011efficient} for fine-grained adjustment. The process is illustrated in Fig. \ref{fig:short-a} and Fig. \ref{fig:short-b}. 

Directly inputting all the labels from the dataset into the VLM could introduce unnecessary noise into the segmentation process. In the first step, we prompt GPT4o to filter the possible categories within each image. The prompt for GPT4o is: ``Help me find as much and accurate as possible, categories appear in the image. The the available categories are \{CATEGORIES\}. Double check if there is anything missing. You may output unsure categories and give them lower probability. Output strictly in the format [id1: classname1, id2: classname2, ...], [probability of class1 in percentage, probability of class2 in percentage, ...] without other words.  '' 
\{CATEGORIES\} being a placeholder for all the categories in a dataset. The categories output by GPT4o with higher than 70\% probability are treated as the predicted classes for the image. 
In the second step, we feed an image and a text prompt to the pretrained VLM and extract cross-attention maps. The text prompt is ``A picture of \texttt{[class 1]} \texttt{[class 2]} ... \texttt{[class K]}'', which includes the $K$ GPT4o predicted categories of the image. 
The image is divided into $P\times P$ patches. The text prompt and the image first go through the modality-specific encoders respectively, followed by a cross-attention fusion module. In the cross-attention layers, the text encodings serve as query vectors and the image patche encodings are as key and value vectors. 

We extract an attention map for each text token to the image patches from a particular cross-attention layer and an attention head. Note that different layers and heads lead to drastic performance differences, and the choices are hyperparameters, tuned using the procedure in \S \ref{sec:hyperparamtuning}. We exclude attention maps for the first three tokens ``A'', ``picture'', ``of'', which do not describe semantic classes. If a class name contains two or more tokens, we take the mean attention map. This procedure yields an attention tensor of size $K \times P \times P$.

With the correct layer and attention head, we observe that this attention map can provide passable semantic segmentations. However, it tends to include many patches unrelated to the class being segmented, leading to over-seg\-mentation. Hence, we introduce two refinement steps, GradCAM and Salience DropOut, explained in \S \ref{sec:gradcam} and \S \ref{sec:salience-dropout}. 
After these steps, we acquire aggregate salience maps for every class. Finally, we conduct local polish to the resultant salience maps using Dense CRF, as described in \S \ref{sec:dense-crf}.

\subsection{Map Sharpening with GradCAM}
\label{sec:gradcam}
The off-the-shelf attention maps tend to also cover many patches unrelated to the class name (See Fig. \ref{fig:attention} (b)). 
Prior works \cite{ALBEF,Tiong-PnPVQA-2022} further sharpen the attention maps and focus them on class-discriminative regions using a variant of GradCAM \cite{GradCAM2017}, which is originally proposed for convolutional networks, but applicable to attention maps. It is worth noting that \cite{ALBEF,Tiong-PnPVQA-2022} use the technique for purposes other than semantic segmentation. 

The GradCAM method requires a gradient. Here we leverage the image-text matching (ITM) loss, which trains the VLM to classify if an image-text pair match each other or not. Computing the ITM loss requires a label. We use ``matching'' (as opposed to ``not matching'') as the label and compute the gradient of the loss with respect to the attention score. This is equivalent to asking: {\em which attention scores contribute the most to the decision that the image-text pair is matching?}
Formally, we denote a $P \times P$ attention map for class $k$ as $M^{(k)}$ and the ITM loss as $\mathcal{L}_{\text{ITM}}$. The GradCAM class salience map is computed as 
\begin{equation}
    \label{cross}
    \widetilde{M}^{(k)} = \max\left(0, \frac{\partial \mathcal{L}_{\text{ITM}}}{\partial M^{(k)}}\right) \otimes M^{(k)},
\end{equation}
\sloppy{where $\otimes$ denote the component-wise multiplication and $\max(\cdot)$ is also applied component-wise. }


\subsection{Salience DropOut}
\label{sec:salience-dropout}


As illustrated in Fig. \ref{fig:attention} (c), segmentations generated by the GradCAM-style re-weighting of cross-attention are often narrowly focused on the most discriminative regions for a given class. However, the less discriminative  regions are still important for the completeness of masks. To compel the VLM to attend to these regions, we propose an iterative technique called Salience DropOut.

We sum up the salience maps $\widetilde{M}^{(k)}$ over the predicted classes $K$, yielding a class-agnostic salience map $U^{(t)}$ for the $t^\text{th}$ Salience DropOut iteration,
$ U^{(t)} = \sum^K_{k=1} \widetilde{M}^{(k, t)}, $
where $\widetilde{M}^{(k, t)}$ is the GradCAM salience map for class $k$ after the $t^\text{th}$ iteration. 

Next, we zero out 10 image patches with the highest values in $U^{(t)}$.
On the remaining image, we compute the GradCAM salience maps $\widetilde{M}^{(k, t+1)}$ and their sum, $U^{(t+1)}$. Any image patch previously zeroed out will always receive zero salience in later iterations. 

Formally, the set of remaining image patches $S^{(t)} \subseteq \{1, \dots, P\}^2$ after the $t^\text{th}$ dropout iteration is defined as

\begin{align}
    S^{(t)} & = S^{(t-1)} \setminus \{ (i,j)~|~ U^{(t)}_{ij} \ge \eta \}, \\
    \eta & = \texttt{median}\left(\{U^{(t)}_{ij}~|~(i,j) \in S^{(t-1)} \} \right),
\end{align}

\begin{align}
    S^{(t)} & = S^{(t-1)} \setminus \{ (i,j)~|~ U^{(t)}_{ij} \in A \}, \\
    A & = \texttt{TOP10}\left(\{U^{(t)}_{ij}~|~(i,j) \in S^{(t-1)} \} \right),
\end{align}

where $U^{(t)}_{i,j}$ denotes the aggregate salience value of the image patch at row $i$ and column $j$. Additionally, note that the input to the first dropout iteration $S^{0} = \{1, \dots, P\}^2$ is the set of all image patches.

We stop at the third rounds of dropout. The final output for each class $k$ is the sum over all salience maps across the three dropout iterations, $\hat{M}^{(k)} = \sum_{t=1}^4  \widetilde{M}^{(k, t)}$.

\subsection{Gaussian Blur and Dense CRF}
\label{sec:dense-crf}

The salience dropout procedure generates $k$ continuous-valued salience maps,  one per object class. To filter out small random noise in the salience values, we subsequent apply a straightforward thresholding operation at a predefined value $T$ on the salience values and obtain binary segmentation masks. Nonetheless, the hard thresholding creates jagged segmentations with sharp edges that often do not coincide with object boundaries. One common strategy in zero-shot segmentation is to apply Dense Conditional Random Field (CRF) \cite{krahenbuhl2011efficient}, which makes fine-grained adjustments to the estimated masks by enforcing consistency between nearby image pixels with similar colors.

However, we find that the hard 0/1 labels in the binary masks do not work well as pixel unary potentials for Dense CRF. Hence, we smooth them using a Gaussian kernel with a preset variance $\sigma$, which results in a better initialization for unary terms and accounts for uncertainty of exact segmentation boundary along the patch boundaries.



\subsection{Hyperparameter Tuning}
\label{sec:hyperparamtuning}

Three hyperparameters in PnP-OVSS have the strongest influence on the result, the cross-attention layer $L$, the attention head $H$, and the binary threshold $T$. Traditionally, tuning these hyperparameters requires a validation set with pixel-level labels. However, since our goal is to perform zero-shot open-vocabulary semantic segmentation, this requirement could potentially limit the applicability of the technique. Instead, we propose a weakly supervised reward function for hyperparameter tuning, which only requires a set of images and the class names appearing in each image. 

The reward for an image $I$ is calculated as follows. We start with a set of classes present in the image, denoted as $\mathcal{K}(I)$. For each class $k \in \mathcal{K}(I)$, we obtain a segmentation mask $M^{(k)}$ (which can be the GradCAM mask, the Salience DropOut mask, or the Dense CRF mask). Next, we apply the mask to the image $I$ and input the extracted regions $M^{(k)} \otimes I$  into a pretrained neural network $f$, which takes an image and a textual class name as input and produces a similarity score.
We calculate the normalized probability that the masked image $M^{(k)} \otimes I$ belongs to the ground-truth class $k$ and contrast with a completely black image $\bm{0}$. 
\begin{align}
    \text{Reward} & = \sum_{k\in \mathcal{K}(I)} \mathds{1} [Pr(M^{(k)} \otimes I, k) > Pr(\bm{0},k) ], \\
    Pr(I,k) & = \frac{\exp(f(I, k))}{ \sum_{k^{\prime}\in \mathcal{K}(I)} \exp(f(I, k^{\prime}))},
\end{align}
where $\mathds{1}(\cdot)$ is the indicator function. 
Intuitively, a reward of $1$ is assigned if and only if the image features pulled with the estimated mask for class $k$ bears higher similarity to the class name of $k$ than a black image (which can be interpreted as the prior probability of class $k$).

We sum up the reward for all validation images as the total reward. The best hyperparameters, including the cross-attention layer, the attention head, the threshold $T$, and the variance in the Gausian blur kernel, are determined using a simple random search.




\section{Experiments}
\label{sec:experiment}

\subsection{Datasets and Implementation Details}

Following the previous work for zero-shot semantic segmentation, we adopt validation sets of Pascal-VOC 2012 \cite{everingham2010pascal}, Pascal Context \cite{mottaghi2014role}, COCO Object \cite{lin2014microsoft}, COCO Stuff \cite{caesar2018coco}, and ADE20K \cite{zhou2017scene} that contain, respectively, 20 object classes, 59 object and stuff classes, 80 object classes, 171 object and stuff classes, and 150 object and stuff classes to evaluate our framework. To verify its versatility, we apply \shortname to two high-performance VLMs, BLIP \cite{li2022blip} and BridgeTower \cite{xu2022bridgetower}, which have the ITM loss and text-to-image cross-attention.  More details are in the supplementary material. 

\begin{table}
  \centering
  \small
  \begin{tabular}{@{}lcccc@{}}
    \toprule

    Hyperparameters &  Start &  End & Step & Solution\\
    \midrule
    {\textbf{\textit{BLIP}}} \\
    Layer & 1 & 12 & 1 & 8\\
    Head & 1 & 12 & 1 & 10\\
    Attention Threshold & 0.05 & 0.5 & 0.1 & 0.15 \\
    \midrule
    {\textbf{\textit{BridgeTower}}} \\
    Layer & 1 & 6 & 1 & 2\\
    Head & 1 & 16 & 1 & 8\\
    Attention Threshold & 0.05 & 0.5 & 0.1 & 0.15\\
    \midrule
    {\textbf{\textit{Gaussian Blur}}} \\
    Standard Deviation & 0.01  & 0.11 & 0.02 & 0.05\\
    
   \bottomrule
  \end{tabular}
  \caption{Search space for hyperparameters}
  \label{tab:searchspace}
\end{table}

\begin{table*}[t]
  \centering
  \small
  \begin{tabular}{@{}lcccccccc@{}}
    \toprule
    \makecell{Method} & \makecell{Finetuning\\VLMs}  & \makecell{HT on Dense\\Labels } &  \makecell{Short-side \\ Resolution} & \makecell{Pascal\\VOC-20} & \makecell{Pascal\\Context-59} & \makecell{COCO\\Object-80} & \makecell{COCO\\Stuff-171}& \makecell{ADE\\20K-150} \\

    \midrule
    \multicolumn{8}{c}{\Gape[0.05cm][0.2cm]{\textit{Group 1: Methods that require weakly supervised finetuning on image-text data}}} \\
    ViL-Seg$^\dagger$ \cite{liu2022open}   & \checkmark & \checkmark& - &37.3  & 18.9 & - & 18.0  & -  \\
    CLIPpy \cite{ranasinghe2023perceptual}  & \checkmark & \checkmark & 224 & 52.2  & - & \textbf{32.0} &  25.5$^\bigstar$ &13.5  \\    
    SegClip \cite{luo2023segclip}  & \checkmark & \checkmark & 224 & 52.6  & 24.7 & 26.5 & - & -\\
    
    GroupVit (by \cite{ranasinghe2023perceptual})  & \checkmark & \checkmark& 224 & 28.1 & 14.8 & 12.9 & - & 6.2\\
    GroupVit (by \cite{cha2023learning})  & \checkmark & \checkmark& 448 & 50.4 & 18.7 & 27.5 & 15.3 & 9.2 \\
    GroupVit \cite{xu2022groupvit} & \checkmark & \checkmark & 448 &52.3 & 22.4 & 24.3 & - & - \\
    ViewCo \cite{ren2023viewco} & \checkmark & \checkmark & 448 & 52.4  & 23.0 & 23.5 & - & -  \\

    OVSegmentor \cite{xu2023learning}   & \checkmark & \checkmark & 448 & 53.8  & 20.4 & 25.1 &  - & 5.6 \\
    
    TCL \cite{cha2023learning} +PAMR \cite{araslanov2020single}& \checkmark & \checkmark & 448  & \underline{55.0}  & \underline{30.4} & \underline{31.6} &  22.4  & \underline{17.1}\\
    PACL \cite{mukhoti2023open}  & \checkmark & \checkmark & 224$\times$4 & \textbf{72.3}  & \textbf{50.1} & \textbf{-} & \textbf{38.8} & \textbf{31.4 }  \\
    \midrule
    \multicolumn{8}{c}{\Gape[0.05cm][0.2cm]{\textit{Group 2: Methods that require finetuning but not real image-text data}}}\\

    MaskClip w/ ST \cite{zhou2022extract}  & \checkmark & \checkmark& 336 & - & \textbf{31.1} & - & \textbf{18.0} & -\\
    MaskClip w/ ST (by \cite{cha2023learning})  & \checkmark & \checkmark& 448 & 38.8 & 23.6 & 20.6 & 16.4 & 9.8\\
    ZeroSeg* \cite{chen2023exploring}  & \checkmark & \checkmark & 448 & 37.3 & 19.7 & 17.8 & - \\
    ZeroSeg \cite{chen2023exploring}  & \checkmark & \checkmark & 448 & \textbf{40.8} & 20.4 & \textbf{20.2} & - & -\\
    \midrule
    \multicolumn{8}{c}{\Gape[0.05cm][0.2cm]{\textit{Group 3: Methods that require no finetuning}}}\\
    MaskClip (by \cite{ranasinghe2023perceptual}) & $\times$ & $\times$ & 224  & 22.1 & - & 13.8 & 8.1 & 6.8\\
    MaskClip \cite{zhou2022extract} & $\times$ & $\times$ & 336  & - & 25.5 & - & 14.6 & -\\
    Reco \cite{shin2022reco} & $\times$ & $\times$ & 320  & - & \textbf{27.2} & - & \textbf{27.2}$^\bigstar$ & - \\
    Reco (by \cite{cha2023learning}) & $\times$ & $\times$ & 448  & \textbf{25.1} & 19.9 & \textbf{15.7} & 14.8 & 11.2\\
    \midrule
    \midrule
    \multicolumn{8}{c}{\Gape[0.05cm][0.2cm]{\textit{PnP-OVSS with different VLMs}}}\\

    BLIP$_{\text{Flickr}}$ & $\times$ & $\times$ & 224 & 46.3& 27.5 & 32.3 & 17.3 & 13.6 \\

    BLIP$_{\text{Flickr}}$ & $\times$ & $\times$  & 336 & \textbf{51.3} & \textbf{28.0} & \textbf{36.2} & \textbf{17.9} & \textbf{14.2} \\

    BridgeTower & $\times$ & $\times$ & 336 & {42.4} & {25.3}  &  {30.4} & {15.7} & 14.8 \\
   \bottomrule
  \end{tabular}
  \caption{Zero-shot semantic segmentation performance in mIoU. Group 3 contains the most similar baselines that serve as fair comparisons to \shortname{}. Groups 1 and 2 benefit from additional training, extra image-text data, and hyperparameter tuning on dense labels. We use the word ``by'' followed by a paper citation to indicate results of the same technique reported by different papers. $^\bigstar$~CLIPpy tests on 133 categories of COCO Stuff and Reco tests on 27 super categories while we test all 171 classes of COCO Stuff.
  ViL-Seg$^\dagger$ is tested on subset of classes on the three datasets, as detailed in the supplementary. 
  }
  \label{tab:main result}
\end{table*}

\begin{table*}
  \centering
  \small
  \begin{tabular}{@{}lccccccc@{}}
    \toprule
    
    Method & \makecell{Dense\\Labels} & \makecell{HT on Dense\\Labels} & \makecell{Pascal\\VOC-20} & \makecell{Pascal\\Context-59} & \makecell{COCO\\Stuff-171}\\
    \midrule

    PnP-OVSS + BLIP$_{\text{Flickr}}$ (Ours)& $\times$& $\times$ & 55.6 & 38.9 & 24.6\\ 
    \midrule
    \midrule
    SPNet+ST \cite{xian2019semantic} & \checkmark  & \checkmark & 25.8 & - & 26.9\\
    ZS3Net+ST \cite{bucher2019zero}  & \checkmark  & \checkmark & 21.2 & 20.7 & 10.6\\
    CaGNet+ST \cite{gu2020context}  &\checkmark  & \checkmark & 30.3 & - & 13.4 \\
    STRICT \cite{pastore2021closer}  &\checkmark  & \checkmark & 35.6 & - & 30.3 \\
    LSeg \cite{li2022languagedriven}  &\checkmark  & \checkmark & 41.0 & - & - \\
    SimBase \cite{xu2022simple}  &\checkmark  & \checkmark & 72.5 & - & 36.3 \\
    MaskCLIP+ w/ ST \cite{zhou2022extract}  &\checkmark  & \checkmark & 86.1 & 66.7 & 54.7 \\

   \bottomrule
  \end{tabular}
  \caption{Comparison of zero-shot semantic segmentation performance on unseen categories with methods trained with dense annotation.}
  \label{tab:main result2}
\end{table*}

\vspace{0.1in}
\noindent \textbf{Hyperparameter tuning.} We use CLIP VIT-L/14 to calculate the reward. The input resolution is 336$\times$336.
The search spaces of the random search and results are shown in Tab. \ref{tab:searchspace}. For computation efficiency, we tune the layer, the head, and the threshold with GradCAM masks. With the first three hyperparameters fixed, we tune the Gaussian variance on masks before Dense CRF. We directly adopt Dense CRF hyperparameters from CutLER \cite{wang2023cut} without tuning. 

\vspace{0.1in}
\noindent \textbf{Baselines.} We adopt recent papers on zero-shot open-vocabulary semantic segmentation as baselines. The only baselines strictly comparable to our work are MaskClip \cite{zhou2022extract} and Reco \cite{shin2022reco}, which do not perform any finetuning on pretrained VLMs and do not perform hyperparameter tuning on dense annotations. We call these Group 3. To further expand our scope, we also include two other groups of baselines. 
Group 2 finetunes VLMs but does not require image-text pairs, including MaskClip with self-training (ST) \cite{zhou2022extract} and ZeroSeg \cite{chen2023exploring}. We include ZeroSeg, trained with ImageNet1K, and ZeroSeg*, trained with CC3M+COCO.
Group 1 contains baselines that require training on image-text pairs.
%
For completeness, we also compare against supervised techniques since 2019.
    To maintain the zero-shot setting, we test them on classes not observed during training. For more details, we refer readers to the the supplementary material.

\begin{table}
  \centering
  \small
  \begin{tabular}{@{}lcc@{}}
    \toprule

    Ablated Model &  \makecell{Pascal\\Context} & \makecell{COCO\\Stuff}\\
    \midrule

    BLIP$_{\text{Flickr}}$ & 18.2 & 13.1 \\
     \;\; + GradCam &20.2 & 14.4\\
     \;\;\;\; + Drop 1  & 21.6 & 14.9 \\
     \;\;\;\; + Drop 2  &22.4 & 15.2 \\
     \;\;\;\; + Drop 3  &22.9& 15.3 \\
     \;\;\;\; + Drop 3 + Blur & 26.1 & 17.1 \\
     \;\;\;\; + Drop 3 + Dense CRF  & 25.1 & 16.0\\
    \;\;\;\; + Drop 3 + Blur + Dense CRF & 28.9 & 17.9\\
    
   \bottomrule
  \end{tabular}
  \caption{An ablation study of \shortname{} + BLIP$_{\text{Flickr}}$ with resolution 336 on Pascal Context and COCO Stuff.}
  \label{tab:ablation}
\end{table}

\subsection{Main Results}
We show the main results in Tab. \ref{tab:main result}. As input resolution may influence results and cause unfair comparisons, we label the resolution used by each method. PACL~\cite{mukhoti2023open} uses an 224 resolution but changes the stride for image patchification from 16 to 4, hence introducing overlapping patches. 

PnP-OVSS exhibits excellent performance. Comparing with MaskClip \cite{zhou2022extract} and Reco \cite{shin2022reco}, the two methods that require no additional training and ground truth for hyperparameter tuning, on an equal-resolution basis, we attain +26.2\% mIoU on Pascal VOC, +0.8\% mIoU on Pascal Context,
+20.5\% mIoU on COCO Object, +3.1\% mIoU COCO Stuff and +3.0\% mIoU ADE-20K.
Further, PnP-OVSS surpasses all baselines except MaskClip with ST in Group 2.

When compared to Group 1, \shortname{} still outperforms most. On the Pascal datasets, under equal or smaller resolutions, PnP-OVSS + BLIP$_{\text{Flickr}}$
outperforms 3 out of 10 baselines on Pascal VOC, and 7 out of 9 baselines on Pascal Context. 
For the COCO datasets, under equal resolutions, PnP-OVSS + BLIP$_{\text{Flickr}}$ beats every baselines on COCO Object, 4 out of 6 baselines on ADE20K . 

When applied to BridgeTower \cite{xu2022bridgetower}, \shortname{} still surpasses all methods in Group 3 by 17.3\% on Pascal VOC, 14.7\% on COCO Object, 0.9\% on COCO Stuff,  and 3.6\% on ADE-20K. This showcases the plug-and-play ability of \shortname{}, which excels with different base networks.

We report comparisons against supervised methods in Tab \ref{tab:main result2}. PnP-OVSS + BLIP$_{\text{Flickr}}$ outperforms 5 out of 7 methods on Pascal VOC, 1 out of 2 methods on Pascal Context and 2 out of 6 methods on COCO stuff. As these baselines benefit from dense supervision, the results further demonstrate the strengths of PnP-OVSS.






\subsection{Ablation Study}

We perform gradual ablation of the components of \shortname{} on BLIP$_{\text{Flickr}}$ and report the results in Tab. \ref{tab:ablation}. Each component, including GradCAM, all Salience DropOut iterations, Gaussian blur, and Dense CRF, contribute positively to the final performance. In particular, the first iteration of Salience DropOut has larger impact (+1.4/0.5) than the second iteration (+0.8/0.3), which in turn is more important than the rest.
Interestingly, Gaussian blur by itself attains good performance +3.2/1.9 whereas Dense CRF only works well when combined with blur. Dense CRF alone is worse than Gaussian blur by 1.0/0.9 mIoU on Pascal Context and COCO stuff.

\begin{table}
  \centering
  \small
  \begin{tabular}{@{}lcccccc@{}}
    \toprule
    
    Layer Mean &  1 &  2 & 3& 4 & 5 & 6\\
    \midrule
    mIoU& 18.6  & 16.4  & 12.5 & 21.2 & 20.4 & 19.5\\
    \midrule
    Layer Mean &7&8&9&10&11 &12 \\
    \midrule
    mIoU& 21.5  & \textbf{25.2}  & 23.6 & 23.0 & 11.7 & 0.8\\
    \midrule
    \midrule
    Head in Layer 8  &  1 &  2 & 3& 4 & 5 & 6\\
    \midrule
    mIoU& 10.4  & 2.0  & 3.3 & 29.8 & 26.2 & 6.5\\
    \midrule
    Head in Layer 8 &7 &8 &9 &10 &11 &12 \\
    \midrule
    mIoU & 12.8  & 3.6  & 3.0 & \textbf{32.3} & 12.4 & 6.2\\
    
   \bottomrule
  \end{tabular}
  \caption{Semantic segmentation performance using cross-attention maps averaged across all heads in a layer and separate heads in Layer 8. Results are attained with \shortname+BLIP$_{\text{Flickr}}$ on COCO Object and resolution 224.}
  \label{tab:hyperparameter sensitivity}
\end{table}

\subsection{Hyperparameter Sensitivity}
\vspace{0.05in}
The choice of hyperparameters often exerts substantial influence on segmentation performance. Here we quantitatively examine how the choice of cross-attention layers and attention heads may change the segmentation mIoU on COCO Stuff. Tab. \ref{tab:hyperparameter sensitivity} shows the results obtained from the average cross-attention maps over all heads in each layer and those from different attention heads. 

We make the following observations. First, different layers and heads have drastic performance differences. The best-to-worst difference among all layers is 24.4, and that among heads in Layer 8 is 30.3, underscoring the importance of hyperparameter tuning. 
Second, the random search using the proposed reward function correctly identifies the best layer-head combination
This indicates the effectiveness of our method. 




    

    

\section{Conclusions}

\vspace{0.05in}

We propose \shortname, which extracts the ability of semantic segmentation from opaque VLMs. \shortname{} is simple to use, requires no extra finetuning, and delivers high performance, exceeding not only all baselines that requires no finetuning, but also many of the baselines that require finetuning. Its success hints at a new direction for open-vocabulary segmentation tasks leveraging large VLMs. 


\section{Acknowledgments}
This work has been supported by the Nanyang Associate Professorship and the National Research Foundation Fellowship (NRF-
NRFF13-2021-0006), Singapore, and by the Vector Institute for AI, Canada CIFAR AI Chair, and NSERC Discovery Grant.
Any opinions, findings, conclusions,
or recommendations expressed in this material are those of the authors and do not reflect the views of the funding agencies.

Earlier version of this paper contained an unintentional error stemming from a bug in the code. This version corrects this error, which had to do with filtering of class names. In consultation with CVPR Program Chairs it was suggested errata be submitted as the updated (fixed) code reinforced original findings (albeit with slightly different final numbers).


{
    \small
    \bibliographystyle{ieeenat_fullname}
    \bibliography{main}

\begin{thebibliography}{91}
\providecommand{\natexlab}[1]{#1}
\providecommand{\url}[1]{\texttt{#1}}
\expandafter\ifx\csname urlstyle\endcsname\relax
  \providecommand{\doi}[1]{doi: #1}\else
  \providecommand{\doi}{doi: \begingroup \urlstyle{rm}\Url}\fi

\bibitem[Alayrac et~al.(2022)Alayrac, Donahue, Luc, Miech, Barr, Hasson, Lenc, Mensch, Millican, Reynolds, Ring, Rutherford, Cabi, Han, Gong, Samangooei, Monteiro, Menick, Borgeaud, Brock, Nematzadeh, Sharifzadeh, Bi\'{n}kowski, Barreira, Vinyals, Zisserman, and Simonyan]{flamingo2022}
Jean-Baptiste Alayrac, Jeff Donahue, Pauline Luc, Antoine Miech, Iain Barr, Yana Hasson, Karel Lenc, Arthur Mensch, Katherine Millican, Malcolm Reynolds, Roman Ring, Eliza Rutherford, Serkan Cabi, Tengda Han, Zhitao Gong, Sina Samangooei, Marianne Monteiro, Jacob~L Menick, Sebastian Borgeaud, Andy Brock, Aida Nematzadeh, Sahand Sharifzadeh, Miko\l~aj Bi\'{n}kowski, Ricardo Barreira, Oriol Vinyals, Andrew Zisserman, and Kar\'{e}n Simonyan.
\newblock Flamingo: a visual language model for few-shot learning.
\newblock In \emph{Advances in Neural Information Processing Systems}, 2022.

\bibitem[Araslanov and Roth(2020)]{araslanov2020single}
Nikita Araslanov and Stefan Roth.
\newblock Single-stage semantic segmentation from image labels.
\newblock In \emph{Proceedings of the IEEE/CVF Conference on Computer Vision and Pattern Recognition}, pages 4253--4262, 2020.

\bibitem[Bazzani et~al.(2016)Bazzani, Bergamo, Anguelov, and Torresani]{bazzani2016self}
Loris Bazzani, Alessandra Bergamo, Dragomir Anguelov, and Lorenzo Torresani.
\newblock Self-taught object localization with deep networks.
\newblock In \emph{2016 IEEE winter conference on applications of computer vision (WACV)}, pages 1--9. IEEE, 2016.

\bibitem[Bucher et~al.(2019)Bucher, Vu, Cord, and P{\'e}rez]{bucher2019zero}
Maxime Bucher, Tuan-Hung Vu, Matthieu Cord, and Patrick P{\'e}rez.
\newblock Zero-shot semantic segmentation.
\newblock \emph{Advances in Neural Information Processing Systems}, 32, 2019.

\bibitem[Caesar et~al.(2018)Caesar, Uijlings, and Ferrari]{caesar2018coco}
Holger Caesar, Jasper Uijlings, and Vittorio Ferrari.
\newblock Coco-stuff: Thing and stuff classes in context.
\newblock In \emph{Proceedings of the IEEE conference on computer vision and pattern recognition}, pages 1209--1218, 2018.

\bibitem[Cha et~al.(2023)Cha, Mun, and Roh]{cha2023learning}
Junbum Cha, Jonghwan Mun, and Byungseok Roh.
\newblock Learning to generate text-grounded mask for open-world semantic segmentation from only image-text pairs.
\newblock In \emph{Proceedings of the IEEE/CVF Conference on Computer Vision and Pattern Recognition}, pages 11165--11174, 2023.

\bibitem[Chen et~al.(2022{\natexlab{a}})Chen, Guo, Yi, Li, and Elhoseiny]{chen2022visualgpt}
Jun Chen, Han Guo, Kai Yi, Boyang Li, and Mohamed Elhoseiny.
\newblock Visualgpt: Data-efficient adaptation of pretrained language models for image captioning.
\newblock In \emph{Proceedings of the IEEE/CVF Conference on Computer Vision and Pattern Recognition}, pages 18030--18040, 2022{\natexlab{a}}.

\bibitem[Chen et~al.(2023)Chen, Zhu, Qian, Ghanem, Yan, Zhu, Xiao, Culatana, and Elhoseiny]{chen2023exploring}
Jun Chen, Deyao Zhu, Guocheng Qian, Bernard Ghanem, Zhicheng Yan, Chenchen Zhu, Fanyi Xiao, Sean~Chang Culatana, and Mohamed Elhoseiny.
\newblock Exploring open-vocabulary semantic segmentation from clip vision encoder distillation only.
\newblock In \emph{Proceedings of the IEEE/CVF International Conference on Computer Vision}, pages 699--710, 2023.

\bibitem[Chen et~al.(2022{\natexlab{b}})Chen, Yang, Lai, and Xie]{chen2022self}
Qi Chen, Lingxiao Yang, Jian-Huang Lai, and Xiaohua Xie.
\newblock Self-supervised image-specific prototype exploration for weakly supervised semantic segmentation.
\newblock In \emph{Proceedings of the IEEE/CVF Conference on Computer Vision and Pattern Recognition}, pages 4288--4298, 2022{\natexlab{b}}.

\bibitem[Chen et~al.(2020)Chen, Li, Yu, El~Kholy, Ahmed, Gan, Cheng, and Liu]{UNITER-2020}
Yen-Chun Chen, Linjie Li, Licheng Yu, Ahmed El~Kholy, Faisal Ahmed, Zhe Gan, Yu Cheng, and Jingjing Liu.
\newblock Uniter: Universal image-text representation learning.
\newblock In \emph{Proceedings of the 16th European Conference on Computer Vision}, 2020.

\bibitem[Chen et~al.(2022{\natexlab{c}})Chen, Wang, Wu, Hua, Zhang, and Sun]{chen2022class}
Zhaozheng Chen, Tan Wang, Xiongwei Wu, Xian-Sheng Hua, Hanwang Zhang, and Qianru Sun.
\newblock Class re-activation maps for weakly-supervised semantic segmentation.
\newblock In \emph{Proceedings of the IEEE/CVF Conference on Computer Vision and Pattern Recognition}, pages 969--978, 2022{\natexlab{c}}.

\bibitem[Cho et~al.(2015)Cho, Kwak, Schmid, and Ponce]{cho2015unsupervised}
Minsu Cho, Suha Kwak, Cordelia Schmid, and Jean Ponce.
\newblock Unsupervised object discovery and localization in the wild: Part-based matching with bottom-up region proposals.
\newblock In \emph{Proceedings of the IEEE conference on computer vision and pattern recognition}, pages 1201--1210, 2015.

\bibitem[Cinbis et~al.(2016)Cinbis, Verbeek, and Schmid]{cinbis2016weakly}
Ramazan~Gokberk Cinbis, Jakob Verbeek, and Cordelia Schmid.
\newblock Weakly supervised object localization with multi-fold multiple instance learning.
\newblock \emph{IEEE transactions on pattern analysis and machine intelligence}, 39\penalty0 (1):\penalty0 189--203, 2016.

\bibitem[Darcet et~al.(2023)Darcet, Oquab, Mairal, and Bojanowski]{darcet2023vision}
Timoth{\'e}e Darcet, Maxime Oquab, Julien Mairal, and Piotr Bojanowski.
\newblock Vision transformers need registers.
\newblock \emph{arXiv preprint arXiv:2309.16588}, 2023.

\bibitem[Du et~al.(2022)Du, Fu, Liu, and Wang]{du2022weakly}
Ye Du, Zehua Fu, Qingjie Liu, and Yunhong Wang.
\newblock Weakly supervised semantic segmentation by pixel-to-prototype contrast.
\newblock In \emph{Proceedings of the IEEE/CVF Conference on Computer Vision and Pattern Recognition}, pages 4320--4329, 2022.

\bibitem[Everingham et~al.(2010)Everingham, Van~Gool, Williams, Winn, and Zisserman]{everingham2010pascal}
Mark Everingham, Luc Van~Gool, Christopher~KI Williams, John Winn, and Andrew Zisserman.
\newblock The pascal visual object classes (voc) challenge.
\newblock \emph{International journal of computer vision}, 88:\penalty0 303--338, 2010.

\bibitem[Ghiasi et~al.(2022)Ghiasi, Gu, Cui, and Lin]{ghiasi2022scaling}
Golnaz Ghiasi, Xiuye Gu, Yin Cui, and Tsung-Yi Lin.
\newblock Scaling open-vocabulary image segmentation with image-level labels.
\newblock In \emph{European Conference on Computer Vision}, pages 540--557. Springer, 2022.

\bibitem[Gu et~al.(2020)Gu, Zhou, Niu, Zhao, and Zhang]{gu2020context}
Zhangxuan Gu, Siyuan Zhou, Li Niu, Zihan Zhao, and Liqing Zhang.
\newblock Context-aware feature generation for zero-shot semantic segmentation.
\newblock In \emph{Proceedings of the 28th ACM International Conference on Multimedia}, pages 1921--1929, 2020.

\bibitem[Guo et~al.(2023)Guo, Li, Li, Tiong, Li, Tao, and Hoi]{guo2023images}
Jiaxian Guo, Junnan Li, Dongxu Li, Anthony Meng~Huat Tiong, Boyang Li, Dacheng Tao, and Steven Hoi.
\newblock From images to textual prompts: Zero-shot visual question answering with frozen large language models.
\newblock In \emph{Proceedings of the IEEE/CVF Conference on Computer Vision and Pattern Recognition}, pages 10867--10877, 2023.

\bibitem[Guo et~al.(2018)Guo, Liu, Georgiou, and Lew]{guo2018review}
Yanming Guo, Yu Liu, Theodoros Georgiou, and Michael~S Lew.
\newblock A review of semantic segmentation using deep neural networks.
\newblock \emph{International journal of multimedia information retrieval}, 7:\penalty0 87--93, 2018.

\bibitem[Han et~al.(2023)Han, Liu, Liew, Ding, Liu, Wang, Tang, Yang, Feng, Zhao, et~al.]{han2023global}
Kunyang Han, Yong Liu, Jun~Hao Liew, Henghui Ding, Jiajun Liu, Yitong Wang, Yansong Tang, Yujiu Yang, Jiashi Feng, Yao Zhao, et~al.
\newblock Global knowledge calibration for fast open-vocabulary segmentation.
\newblock In \emph{Proceedings of the IEEE/CVF International Conference on Computer Vision}, pages 797--807, 2023.

\bibitem[Hao et~al.(2020)Hao, Zhou, and Guo]{hao2020brief}
Shijie Hao, Yuan Zhou, and Yanrong Guo.
\newblock A brief survey on semantic segmentation with deep learning.
\newblock \emph{Neurocomputing}, 406:\penalty0 302--321, 2020.

\bibitem[Hou et~al.(2018)Hou, Jiang, Wei, and Cheng]{hou2018self}
Qibin Hou, PengTao Jiang, Yunchao Wei, and Ming-Ming Cheng.
\newblock Self-erasing network for integral object attention.
\newblock \emph{Advances in Neural Information Processing Systems}, 31, 2018.

\bibitem[Jia et~al.(2021)Jia, Yang, Xia, Chen, Parekh, Pham, Le, Sung, Li, and Duerig]{Jia-ALIGN-2021}
Chao Jia, Yinfei Yang, Ye Xia, Yi-Ting Chen, Zarana Parekh, Hieu Pham, Quoc Le, Yun-Hsuan Sung, Zhen Li, and Tom Duerig.
\newblock Scaling up visual and vision-language representation learning with noisy text supervision.
\newblock In \emph{Proceedings of the 38th International Conference on Machine Learning}, 2021.

\bibitem[Jiang et~al.(2022)Jiang, Yang, Hou, and Wei]{jiang2022l2g}
Peng-Tao Jiang, Yuqi Yang, Qibin Hou, and Yunchao Wei.
\newblock L2g: A simple local-to-global knowledge transfer framework for weakly supervised semantic segmentation.
\newblock In \emph{Proceedings of the IEEE/CVF conference on computer vision and pattern recognition}, pages 16886--16896, 2022.

\bibitem[Ke et~al.(2021)Ke, Hwang, and Yu]{ke2021universal}
Tsung-Wei Ke, Jyh-Jing Hwang, and Stella~X Yu.
\newblock Universal weakly supervised segmentation by pixel-to-segment contrastive learning.
\newblock \emph{arXiv preprint arXiv:2105.00957}, 2021.

\bibitem[Kim et~al.(2021)Kim, Son, and Kim]{Wonjae-ViLT-2021}
Wonjae Kim, Bokyung Son, and Ildoo Kim.
\newblock Vilt: Vision-and-language transformer without convolution or region supervision.
\newblock In \emph{Proceedings of the 38th International Conference on Machine Learning}, 2021.

\bibitem[Kolesnikov and Lampert(2016)]{kolesnikov2016seed}
Alexander Kolesnikov and Christoph~H Lampert.
\newblock Seed, expand and constrain: Three principles for weakly-supervised image segmentation.
\newblock In \emph{Computer Vision--ECCV 2016: 14th European Conference, Amsterdam, The Netherlands, October 11--14, 2016, Proceedings, Part IV 14}, pages 695--711. Springer, 2016.

\bibitem[Kr{\"a}henb{\"u}hl and Koltun(2011)]{krahenbuhl2011efficient}
Philipp Kr{\"a}henb{\"u}hl and Vladlen Koltun.
\newblock Efficient inference in fully connected crfs with gaussian edge potentials.
\newblock \emph{Advances in neural information processing systems}, 24, 2011.

\bibitem[Kumar~Singh and Jae~Lee(2017)]{kumar2017hide}
Krishna Kumar~Singh and Yong Jae~Lee.
\newblock Hide-and-seek: Forcing a network to be meticulous for weakly-supervised object and action localization.
\newblock In \emph{Proceedings of the IEEE International Conference on Computer Vision}, pages 3524--3533, 2017.

\bibitem[Lee et~al.(2022)Lee, Oh, Yun, Choe, Kim, and Yoon]{lee2022weakly}
Jungbeom Lee, Seong~Joon Oh, Sangdoo Yun, Junsuk Choe, Eunji Kim, and Sungroh Yoon.
\newblock Weakly supervised semantic segmentation using out-of-distribution data.
\newblock In \emph{Proceedings of the IEEE/CVF Conference on Computer Vision and Pattern Recognition}, pages 16897--16906, 2022.

\bibitem[Li et~al.(2022{\natexlab{a}})Li, Weinberger, Belongie, Koltun, and Ranftl]{li2022languagedriven}
Boyi Li, Kilian~Q. Weinberger, Serge Belongie, Vladlen Koltun, and René Ranftl.
\newblock Language-driven semantic segmentation, 2022{\natexlab{a}}.

\bibitem[Li et~al.(2022{\natexlab{b}})Li, Xu, Tian, Wang, Yan, Bi, Ye, Chen, Xu, Cao, Zhang, Huang, Huang, Zhou, and Si]{li-etal-2022-mplug}
Chenliang Li, Haiyang Xu, Junfeng Tian, Wei Wang, Ming Yan, Bin Bi, Jiabo Ye, He Chen, Guohai Xu, Zheng Cao, Ji Zhang, Songfang Huang, Fei Huang, Jingren Zhou, and Luo Si.
\newblock m{PLUG}: Effective and efficient vision-language learning by cross-modal skip-connections.
\newblock In \emph{Proceedings of the 2022 Conference on Empirical Methods in Natural Language Processing}, 2022{\natexlab{b}}.

\bibitem[Li et~al.(2022{\natexlab{c}})Li, Xu, Tian, Wang, Yan, Bi, Ye, Chen, Xu, Cao, et~al.]{li2022mplug}
Chenliang Li, Haiyang Xu, Junfeng Tian, Wei Wang, Ming Yan, Bin Bi, Jiabo Ye, Hehong Chen, Guohai Xu, Zheng Cao, et~al.
\newblock mplug: Effective and efficient vision-language learning by cross-modal skip-connections.
\newblock \emph{arXiv preprint arXiv:2205.12005}, 2022{\natexlab{c}}.

\bibitem[Li et~al.(2021{\natexlab{a}})Li, Selvaraju, Gotmare, Joty, Xiong, and Hoi]{ALBEF}
Junnan Li, Ramprasaath~R. Selvaraju, Akhilesh~Deepak Gotmare, Shafiq Joty, Caiming Xiong, and Steven Hoi.
\newblock Align before fuse: Vision and language representation learning with momentum distillation.
\newblock In \emph{NeurIPS}, 2021{\natexlab{a}}.

\bibitem[Li et~al.(2022{\natexlab{d}})Li, Jie, Wang, Wei, and Ma]{li2022expansion}
Jinlong Li, Zequn Jie, Xu Wang, Xiaolin Wei, and Lin Ma.
\newblock Expansion and shrinkage of localization for weakly-supervised semantic segmentation.
\newblock \emph{Advances in Neural Information Processing Systems}, 35:\penalty0 16037--16051, 2022{\natexlab{d}}.

\bibitem[Li et~al.(2022{\natexlab{e}})Li, Li, Xiong, and Hoi]{li2022blip}
Junnan Li, Dongxu Li, Caiming Xiong, and Steven Hoi.
\newblock Blip: Bootstrapping language-image pre-training for unified vision-language understanding and generation.
\newblock In \emph{International Conference on Machine Learning}, pages 12888--12900. PMLR, 2022{\natexlab{e}}.

\bibitem[Li et~al.(2023)Li, Li, Savarese, and Hoi]{li2023blip2}
Junnan Li, Dongxu Li, Silvio Savarese, and Steven Hoi.
\newblock Blip-2: Bootstrapping language-image pre-training with frozen image encoders and large language models.
\newblock In \emph{ICML}, 2023.

\bibitem[Li et~al.(2021{\natexlab{b}})Li, Gao, Niu, Xiao, Liu, Liu, Wu, and Wang]{li-etal-2021-unimo}
Wei Li, Can Gao, Guocheng Niu, Xinyan Xiao, Hao Liu, Jiachen Liu, Hua Wu, and Haifeng Wang.
\newblock {UNIMO}: Towards unified-modal understanding and generation via cross-modal contrastive learning.
\newblock In \emph{Proceedings of the 59th Annual Meeting of the Association for Computational Linguistics and the 11th International Joint Conference on Natural Language Processing (Volume 1: Long Papers)}, 2021{\natexlab{b}}.

\bibitem[Li et~al.(2020)Li, Yin, Li, Zhang, Hu, Zhang, Wang, Hu, Dong, Wei, Choi, and Gao]{OSCAR-2020}
Xiujun Li, Xi Yin, Chunyuan Li, Pengchuan Zhang, Xiaowei Hu, Lei Zhang, Lijuan Wang, Houdong Hu, Li Dong, Furu Wei, Yejin Choi, and Jianfeng Gao.
\newblock Oscar: Object-semantics aligned pre-training for vision-language tasks.
\newblock In \emph{ECCV 2020}, 2020.

\bibitem[Liang et~al.(2023)Liang, Wu, Dai, Li, Zhao, Zhang, Zhang, Vajda, and Marculescu]{liang2023open}
Feng Liang, Bichen Wu, Xiaoliang Dai, Kunpeng Li, Yinan Zhao, Hang Zhang, Peizhao Zhang, Peter Vajda, and Diana Marculescu.
\newblock Open-vocabulary semantic segmentation with mask-adapted clip.
\newblock In \emph{Proceedings of the IEEE/CVF Conference on Computer Vision and Pattern Recognition}, pages 7061--7070, 2023.

\bibitem[Lin et~al.(2014)Lin, Maire, Belongie, Hays, Perona, Ramanan, Doll{\'a}r, and Zitnick]{lin2014microsoft}
Tsung-Yi Lin, Michael Maire, Serge Belongie, James Hays, Pietro Perona, Deva Ramanan, Piotr Doll{\'a}r, and C~Lawrence Zitnick.
\newblock Microsoft coco: Common objects in context.
\newblock In \emph{Computer Vision--ECCV 2014: 13th European Conference, Zurich, Switzerland, September 6-12, 2014, Proceedings, Part V 13}, pages 740--755. Springer, 2014.

\bibitem[Liu et~al.(2022)Liu, Wen, Han, Xu, Xu, and Liang]{liu2022open}
Quande Liu, Youpeng Wen, Jianhua Han, Chunjing Xu, Hang Xu, and Xiaodan Liang.
\newblock Open-world semantic segmentation via contrasting and clustering vision-language embedding.
\newblock In \emph{European Conference on Computer Vision}, pages 275--292. Springer, 2022.

\bibitem[Luo et~al.(2023)Luo, Bao, Wu, He, and Li]{luo2023segclip}
Huaishao Luo, Junwei Bao, Youzheng Wu, Xiaodong He, and Tianrui Li.
\newblock Segclip: Patch aggregation with learnable centers for open-vocabulary semantic segmentation.
\newblock In \emph{International Conference on Machine Learning}, pages 23033--23044. PMLR, 2023.

\bibitem[Mottaghi et~al.(2014)Mottaghi, Chen, Liu, Cho, Lee, Fidler, Urtasun, and Yuille]{mottaghi2014role}
Roozbeh Mottaghi, Xianjie Chen, Xiaobai Liu, Nam-Gyu Cho, Seong-Whan Lee, Sanja Fidler, Raquel Urtasun, and Alan Yuille.
\newblock The role of context for object detection and semantic segmentation in the wild.
\newblock In \emph{Proceedings of the IEEE conference on computer vision and pattern recognition}, pages 891--898, 2014.

\bibitem[Mukhoti et~al.(2023)Mukhoti, Lin, Poursaeed, Wang, Shah, Torr, and Lim]{mukhoti2023open}
Jishnu Mukhoti, Tsung-Yu Lin, Omid Poursaeed, Rui Wang, Ashish Shah, Philip~HS Torr, and Ser-Nam Lim.
\newblock Open vocabulary semantic segmentation with patch aligned contrastive learning.
\newblock In \emph{Proceedings of the IEEE/CVF Conference on Computer Vision and Pattern Recognition}, pages 19413--19423, 2023.

\bibitem[Pastore et~al.(2021)Pastore, Cermelli, Xian, Mancini, Akata, and Caputo]{pastore2021closer}
Giuseppe Pastore, Fabio Cermelli, Yongqin Xian, Massimiliano Mancini, Zeynep Akata, and Barbara Caputo.
\newblock A closer look at self-training for zero-label semantic segmentation.
\newblock In \emph{Proceedings of the IEEE/CVF Conference on Computer Vision and Pattern Recognition}, pages 2693--2702, 2021.

\bibitem[Qin et~al.(2023)Qin, Wu, Yan, Li, Yuxi, Xiao, Wang, Wang, Wen, Pan, et~al.]{qin2023freeseg}
Jie Qin, Jie Wu, Pengxiang Yan, Ming Li, Ren Yuxi, Xuefeng Xiao, Yitong Wang, Rui Wang, Shilei Wen, Xin Pan, et~al.
\newblock Freeseg: Unified, universal and open-vocabulary image segmentation.
\newblock In \emph{Proceedings of the IEEE/CVF Conference on Computer Vision and Pattern Recognition}, pages 19446--19455, 2023.

\bibitem[Radford et~al.(2021{\natexlab{a}})Radford, Kim, Hallacy, Ramesh, Goh, Agarwal, Sastry, Askell, Mishkin, Clark, Krueger, and Sutskever]{radford-clip-2021}
Alec Radford, Jong~Wook Kim, Chris Hallacy, Aditya Ramesh, Gabriel Goh, Sandhini Agarwal, Girish Sastry, Amanda Askell, Pamela Mishkin, Jack Clark, Gretchen Krueger, and Ilya Sutskever.
\newblock Learning transferable visual models from natural language supervision.
\newblock In \emph{Proceedings of the 38th International Conference on Machine Learning}, 2021{\natexlab{a}}.

\bibitem[Radford et~al.(2021{\natexlab{b}})Radford, Kim, Hallacy, Ramesh, Goh, Agarwal, Sastry, Askell, Mishkin, Clark, et~al.]{radford2021learning}
Alec Radford, Jong~Wook Kim, Chris Hallacy, Aditya Ramesh, Gabriel Goh, Sandhini Agarwal, Girish Sastry, Amanda Askell, Pamela Mishkin, Jack Clark, et~al.
\newblock Learning transferable visual models from natural language supervision.
\newblock In \emph{International conference on machine learning}, pages 8748--8763. PMLR, 2021{\natexlab{b}}.

\bibitem[Ranasinghe et~al.(2023)Ranasinghe, McKinzie, Ravi, Yang, Toshev, and Shlens]{ranasinghe2023perceptual}
Kanchana Ranasinghe, Brandon McKinzie, Sachin Ravi, Yinfei Yang, Alexander Toshev, and Jonathon Shlens.
\newblock Perceptual grouping in contrastive vision-language models.
\newblock In \emph{Proceedings of the IEEE/CVF International Conference on Computer Vision}, pages 5571--5584, 2023.

\bibitem[Ren et~al.(2023)Ren, Li, Xu, Zhu, Wang, Liu, Chang, and Liang]{ren2023viewco}
Pengzhen Ren, Changlin Li, Hang Xu, Yi Zhu, Guangrun Wang, Jianzhuang Liu, Xiaojun Chang, and Xiaodan Liang.
\newblock Viewco: Discovering text-supervised segmentation masks via multi-view semantic consistency.
\newblock \emph{arXiv preprint arXiv:2302.10307}, 2023.

\bibitem[Ru et~al.(2022)Ru, Zhan, Yu, and Du]{ru2022learning}
Lixiang Ru, Yibing Zhan, Baosheng Yu, and Bo Du.
\newblock Learning affinity from attention: End-to-end weakly-supervised semantic segmentation with transformers.
\newblock In \emph{Proceedings of the IEEE/CVF Conference on Computer Vision and Pattern Recognition}, pages 16846--16855, 2022.

\bibitem[Ru et~al.(2023{\natexlab{a}})Ru, Zheng, Zhan, and Du]{ru2023toco}
Lixiang Ru, Heliang Zheng, Yibing Zhan, and Bo Du.
\newblock Token contrast for weakly-supervised semantic segmentation.
\newblock In \emph{Proceedings of the IEEE/CVF Conference on Computer Vision and Pattern Recognition}, pages 3093--3102, 2023{\natexlab{a}}.

\bibitem[Ru et~al.(2023{\natexlab{b}})Ru, Zheng, Zhan, and Du]{ru2023token}
Lixiang Ru, Heliang Zheng, Yibing Zhan, and Bo Du.
\newblock Token contrast for weakly-supervised semantic segmentation.
\newblock In \emph{Proceedings of the IEEE/CVF Conference on Computer Vision and Pattern Recognition}, pages 3093--3102, 2023{\natexlab{b}}.

\bibitem[Rubinstein et~al.(2013)Rubinstein, Joulin, Kopf, and Liu]{rubinstein2013unsupervised}
Michael Rubinstein, Armand Joulin, Johannes Kopf, and Ce Liu.
\newblock Unsupervised joint object discovery and segmentation in internet images.
\newblock In \emph{Proceedings of the IEEE conference on computer vision and pattern recognition}, pages 1939--1946, 2013.

\bibitem[Selvaraju et~al.(2017)Selvaraju, Cogswell, Das, Vedantam, Parikh, and Batra]{GradCAM2017}
Ramprasaath~R. Selvaraju, Michael Cogswell, Abhishek Das, Ramakrishna Vedantam, Devi Parikh, and Dhruv Batra.
\newblock Grad-cam: Visual explanations from deep networks via gradient-based localization.
\newblock In \emph{Proceedings of the IEEE/CVF International Conference on Computer Vision}, 2017.

\bibitem[Shin et~al.(2022)Shin, Xie, and Albanie]{shin2022reco}
Gyungin Shin, Weidi Xie, and Samuel Albanie.
\newblock Reco: Retrieve and co-segment for zero-shot transfer.
\newblock \emph{Advances in Neural Information Processing Systems}, 35:\penalty0 33754--33767, 2022.

\bibitem[Sim{\'e}oni et~al.(2021)Sim{\'e}oni, Puy, Vo, Roburin, Gidaris, Bursuc, P{\'e}rez, Marlet, and Ponce]{simeoni2021localizing}
Oriane Sim{\'e}oni, Gilles Puy, Huy~V Vo, Simon Roburin, Spyros Gidaris, Andrei Bursuc, Patrick P{\'e}rez, Renaud Marlet, and Jean Ponce.
\newblock Localizing objects with self-supervised transformers and no labels.
\newblock \emph{arXiv preprint arXiv:2109.14279}, 2021.

\bibitem[{Simon Blanke}(since 2020)]{gfo2020}
{Simon Blanke}.
\newblock {Gradient-Free-Optimizers}: Simple and reliable optimization with local, global, population-based and sequential techniques in numerical search spaces.
\newblock \url{https://github.com/SimonBlanke}, since 2020.

\bibitem[Singh et~al.(2022)Singh, Hu, Goswami, Couairon, Galuba, Rohrbach, and Kiela]{Singh_FLAVA_2022}
Amanpreet Singh, Ronghang Hu, Vedanuj Goswami, Guillaume Couairon, Wojciech Galuba, Marcus Rohrbach, and Douwe Kiela.
\newblock Flava: A foundational language and vision alignment model.
\newblock In \emph{Proceedings of the IEEE/CVF Conference on Computer Vision and Pattern Recognition (CVPR)}, pages 15638--15650, 2022.

\bibitem[Tang et~al.(2018)Tang, Perazzi, Djelouah, Ben~Ayed, Schroers, and Boykov]{tang2018regularized}
Meng Tang, Federico Perazzi, Abdelaziz Djelouah, Ismail Ben~Ayed, Christopher Schroers, and Yuri Boykov.
\newblock On regularized losses for weakly-supervised cnn segmentation.
\newblock In \emph{Proceedings of the European Conference on Computer Vision (ECCV)}, pages 507--522, 2018.

\bibitem[Tiong et~al.(2022)Tiong, Li, Li, Savarese, and Hoi]{Tiong-PnPVQA-2022}
Anthony Meng~Huat Tiong, Junnan Li, Boyang Li, Silvio Savarese, and Steven~C.H. Hoi.
\newblock Plug-and-play vqa: Zero-shot vqa by conjoining large pretrained models with zero training.
\newblock In \emph{Findings of the Conference on Empirical Methods in Natural Language Processing (Findings of EMNLP)}, 2022.

\bibitem[Vo et~al.(2019)Vo, Bach, Cho, Han, LeCun, P{\'e}rez, and Ponce]{vo2019unsupervised}
Huy~V Vo, Francis Bach, Minsu Cho, Kai Han, Yann LeCun, Patrick P{\'e}rez, and Jean Ponce.
\newblock Unsupervised image matching and object discovery as optimization.
\newblock In \emph{Proceedings of the IEEE/CVF Conference on Computer Vision and Pattern Recognition}, pages 8287--8296, 2019.

\bibitem[Vo et~al.(2021)Vo, Sizikova, Schmid, P{\'e}rez, and Ponce]{vo2021large}
Van~Huy Vo, Elena Sizikova, Cordelia Schmid, Patrick P{\'e}rez, and Jean Ponce.
\newblock Large-scale unsupervised object discovery.
\newblock \emph{Advances in Neural Information Processing Systems}, 34:\penalty0 16764--16778, 2021.

\bibitem[Wang et~al.(2022{\natexlab{a}})Wang, Sun, and Van~Gool]{wang2022looking}
Wenguan Wang, Guolei Sun, and Luc Van~Gool.
\newblock Looking beyond single images for weakly supervised semantic segmentation learning.
\newblock \emph{IEEE Transactions on Pattern Analysis and Machine Intelligence}, 2022{\natexlab{a}}.

\bibitem[Wang et~al.(2023)Wang, Girdhar, Yu, and Misra]{wang2023cut}
Xudong Wang, Rohit Girdhar, Stella~X Yu, and Ishan Misra.
\newblock Cut and learn for unsupervised object detection and instance segmentation.
\newblock In \emph{Proceedings of the IEEE/CVF Conference on Computer Vision and Pattern Recognition}, pages 3124--3134, 2023.

\bibitem[Wang et~al.(2022{\natexlab{b}})Wang, Shen, Hu, Yuan, Crowley, and Vaufreydaz]{wang2022self}
Yangtao Wang, Xi Shen, Shell~Xu Hu, Yuan Yuan, James~L Crowley, and Dominique Vaufreydaz.
\newblock Self-supervised transformers for unsupervised object discovery using normalized cut.
\newblock In \emph{Proceedings of the IEEE/CVF Conference on Computer Vision and Pattern Recognition}, pages 14543--14553, 2022{\natexlab{b}}.

\bibitem[Wei et~al.(2019)Wei, Zhang, Wu, Shen, and Zhou]{wei2019unsupervised}
Xiu-Shen Wei, Chen-Lin Zhang, Jianxin Wu, Chunhua Shen, and Zhi-Hua Zhou.
\newblock Unsupervised object discovery and co-localization by deep descriptor transformation.
\newblock \emph{Pattern Recognition}, 88:\penalty0 113--126, 2019.

\bibitem[Wei et~al.(2017)Wei, Feng, Liang, Cheng, Zhao, and Yan]{wei2017object}
Yunchao Wei, Jiashi Feng, Xiaodan Liang, Ming-Ming Cheng, Yao Zhao, and Shuicheng Yan.
\newblock Object region mining with adversarial erasing: A simple classification to semantic segmentation approach.
\newblock In \emph{Proceedings of the IEEE conference on computer vision and pattern recognition}, pages 1568--1576, 2017.

\bibitem[Xenos et~al.(2023)Xenos, Stafylakis, Patras, and Tzimiropoulos]{xenos2023simple}
Alexandros Xenos, Themos Stafylakis, Ioannis Patras, and Georgios Tzimiropoulos.
\newblock A simple baseline for knowledge-based visual question answering.
\newblock In \emph{EMNLP}, 2023.

\bibitem[Xian et~al.(2019)Xian, Choudhury, He, Schiele, and Akata]{xian2019semantic}
Yongqin Xian, Subhabrata Choudhury, Yang He, Bernt Schiele, and Zeynep Akata.
\newblock Semantic projection network for zero-and few-label semantic segmentation.
\newblock In \emph{Proceedings of the IEEE/CVF Conference on Computer Vision and Pattern Recognition}, pages 8256--8265, 2019.

\bibitem[Xu et~al.(2023{\natexlab{a}})Xu, Ye, Yan, Shi, Ye, Xu, Li, Bi, Qian, Wang, et~al.]{xu2023mplug}
Haiyang Xu, Qinghao Ye, Ming Yan, Yaya Shi, Jiabo Ye, Yuanhong Xu, Chenliang Li, Bin Bi, Qi Qian, Wei Wang, et~al.
\newblock mplug-2: A modularized multi-modal foundation model across text, image and video.
\newblock \emph{arXiv preprint arXiv:2302.00402}, 2023{\natexlab{a}}.

\bibitem[Xu et~al.(2022{\natexlab{a}})Xu, De~Mello, Liu, Byeon, Breuel, Kautz, and Wang]{xu2022groupvit}
Jiarui Xu, Shalini De~Mello, Sifei Liu, Wonmin Byeon, Thomas Breuel, Jan Kautz, and Xiaolong Wang.
\newblock Groupvit: Semantic segmentation emerges from text supervision.
\newblock In \emph{Proceedings of the IEEE/CVF Conference on Computer Vision and Pattern Recognition}, pages 18134--18144, 2022{\natexlab{a}}.

\bibitem[Xu et~al.(2023{\natexlab{b}})Xu, Hou, Zhang, Feng, Wang, Qiao, and Xie]{xu2023learning}
Jilan Xu, Junlin Hou, Yuejie Zhang, Rui Feng, Yi Wang, Yu Qiao, and Weidi Xie.
\newblock Learning open-vocabulary semantic segmentation models from natural language supervision.
\newblock In \emph{Proceedings of the IEEE/CVF Conference on Computer Vision and Pattern Recognition}, pages 2935--2944, 2023{\natexlab{b}}.

\bibitem[Xu et~al.(2022{\natexlab{b}})Xu, Ouyang, Bennamoun, Boussaid, and Xu]{xu2022multi}
Lian Xu, Wanli Ouyang, Mohammed Bennamoun, Farid Boussaid, and Dan Xu.
\newblock Multi-class token transformer for weakly supervised semantic segmentation.
\newblock In \emph{Proceedings of the IEEE/CVF Conference on Computer Vision and Pattern Recognition}, pages 4310--4319, 2022{\natexlab{b}}.

\bibitem[Xu et~al.(2022{\natexlab{c}})Xu, Zhang, Wei, Lin, Cao, Hu, and Bai]{xu2022simple}
Mengde Xu, Zheng Zhang, Fangyun Wei, Yutong Lin, Yue Cao, Han Hu, and Xiang Bai.
\newblock A simple baseline for open-vocabulary semantic segmentation with pre-trained vision-language model.
\newblock In \emph{European Conference on Computer Vision}, pages 736--753. Springer, 2022{\natexlab{c}}.

\bibitem[Xu et~al.(2023{\natexlab{c}})Xu, Zhang, Wei, Hu, and Bai]{xu2023side}
Mengde Xu, Zheng Zhang, Fangyun Wei, Han Hu, and Xiang Bai.
\newblock Side adapter network for open-vocabulary semantic segmentation.
\newblock In \emph{Proceedings of the IEEE/CVF Conference on Computer Vision and Pattern Recognition}, pages 2945--2954, 2023{\natexlab{c}}.

\bibitem[Xu et~al.(2022{\natexlab{d}})Xu, Wu, Rosenman, Lal, Che, and Duan]{xu2022bridgetower}
Xiao Xu, Chenfei Wu, Shachar Rosenman, Vasudev Lal, Wanxiang Che, and Nan Duan.
\newblock Bridgetower: Building bridges between encoders in vision-language representation learning.
\newblock \emph{arXiv preprint arXiv:2206.08657}, 2022{\natexlab{d}}.

\bibitem[Yu et~al.(2022)Yu, Wang, Vasudevan, Yeung, Seyedhosseini, and Wu]{yu2022coca}
Jiahui Yu, Zirui Wang, Vijay Vasudevan, Legg Yeung, Mojtaba Seyedhosseini, and Yonghui Wu.
\newblock {CoCa}: Contrastive captioners are image-text foundation models.
\newblock \emph{arXiv Preprint 2205.01917}, 2022.

\bibitem[Zabari and Hoshen(2022)]{zabari2022open}
Nir Zabari and Yedid Hoshen.
\newblock Open-vocabulary semantic segmentation using test-time distillation.
\newblock In \emph{European Conference on Computer Vision}, pages 56--72. Springer, 2022.

\bibitem[Zeng et~al.(2021)Zeng, Zhang, and Li]{zeng2021multi}
Yan Zeng, Xinsong Zhang, and Hang Li.
\newblock Multi-grained vision language pre-training: Aligning texts with visual concepts.
\newblock \emph{arXiv preprint arXiv:2111.08276}, 2021.

\bibitem[Zeng et~al.(2022)Zeng, Zhang, Li, Wang, Zhang, and Zhou]{zeng2022x}
Yan Zeng, Xinsong Zhang, Hang Li, Jiawei Wang, Jipeng Zhang, and Wangchunshu Zhou.
\newblock X$^{2}$-vlm: All-in-one pre-trained model for vision-language tasks.
\newblock \emph{arXiv preprint arXiv:2211.12402}, 2022.

\bibitem[Zhang et~al.(2020)Zhang, Xiao, Wei, Sun, and Huang]{zhang2020reliability}
Bingfeng Zhang, Jimin Xiao, Yunchao Wei, Mingjie Sun, and Kaizhu Huang.
\newblock Reliability does matter: An end-to-end weakly supervised semantic segmentation approach.
\newblock In \emph{Proceedings of the AAAI Conference on Artificial Intelligence}, pages 12765--12772, 2020.

\bibitem[Zhang et~al.(2023)Zhang, Li, Zou, Liu, Li, Yang, and Zhang]{zhang2023simple}
Hao Zhang, Feng Li, Xueyan Zou, Shilong Liu, Chunyuan Li, Jianwei Yang, and Lei Zhang.
\newblock A simple framework for open-vocabulary segmentation and detection.
\newblock In \emph{Proceedings of the IEEE/CVF International Conference on Computer Vision}, pages 1020--1031, 2023.

\bibitem[Zhang et~al.(2021)Zhang, Li, Hu, Yang, Zhang, Wang, Choi, and Gao]{Zhang_VinVL_2021}
Pengchuan Zhang, Xiujun Li, Xiaowei Hu, Jianwei Yang, Lei Zhang, Lijuan Wang, Yejin Choi, and Jianfeng Gao.
\newblock Vinvl: Revisiting visual representations in vision-language models.
\newblock In \emph{Proceedings of the IEEE/CVF Conference on Computer Vision and Pattern Recognition (CVPR)}, pages 5579--5588, 2021.

\bibitem[Zhang et~al.(2018)Zhang, Wei, Feng, Yang, and Huang]{zhang2018adversarial}
Xiaolin Zhang, Yunchao Wei, Jiashi Feng, Yi Yang, and Thomas~S Huang.
\newblock Adversarial complementary learning for weakly supervised object localization.
\newblock In \emph{Proceedings of the IEEE conference on computer vision and pattern recognition}, pages 1325--1334, 2018.

\bibitem[Zhou et~al.(2014)Zhou, Khosla, Lapedriza, Oliva, and Torralba]{zhou2014object}
Bolei Zhou, Aditya Khosla, Agata Lapedriza, Aude Oliva, and Antonio Torralba.
\newblock Object detectors emerge in deep scene cnns.
\newblock \emph{arXiv preprint arXiv:1412.6856}, 2014.

\bibitem[Zhou et~al.(2017)Zhou, Zhao, Puig, Fidler, Barriuso, and Torralba]{zhou2017scene}
Bolei Zhou, Hang Zhao, Xavier Puig, Sanja Fidler, Adela Barriuso, and Antonio Torralba.
\newblock Scene parsing through ade20k dataset.
\newblock In \emph{Proceedings of the IEEE conference on computer vision and pattern recognition}, pages 633--641, 2017.

\bibitem[Zhou et~al.(2022{\natexlab{a}})Zhou, Loy, and Dai]{zhou2022extract}
Chong Zhou, Chen~Change Loy, and Bo Dai.
\newblock Extract free dense labels from clip.
\newblock In \emph{European Conference on Computer Vision}, pages 696--712. Springer, 2022{\natexlab{a}}.

\bibitem[Zhou et~al.(2022{\natexlab{b}})Zhou, Zhang, Zhao, and Li]{zhou2022regional}
Tianfei Zhou, Meijie Zhang, Fang Zhao, and Jianwu Li.
\newblock Regional semantic contrast and aggregation for weakly supervised semantic segmentation.
\newblock In \emph{Proceedings of the IEEE/CVF Conference on Computer Vision and Pattern Recognition}, pages 4299--4309, 2022{\natexlab{b}}.

\end{thebibliography}
}

\clearpage
\setcounter{page}{1}
\maketitlesupplementary

\begin{figure*}[t]
  \centering
    \includegraphics[width=0.95\textwidth]{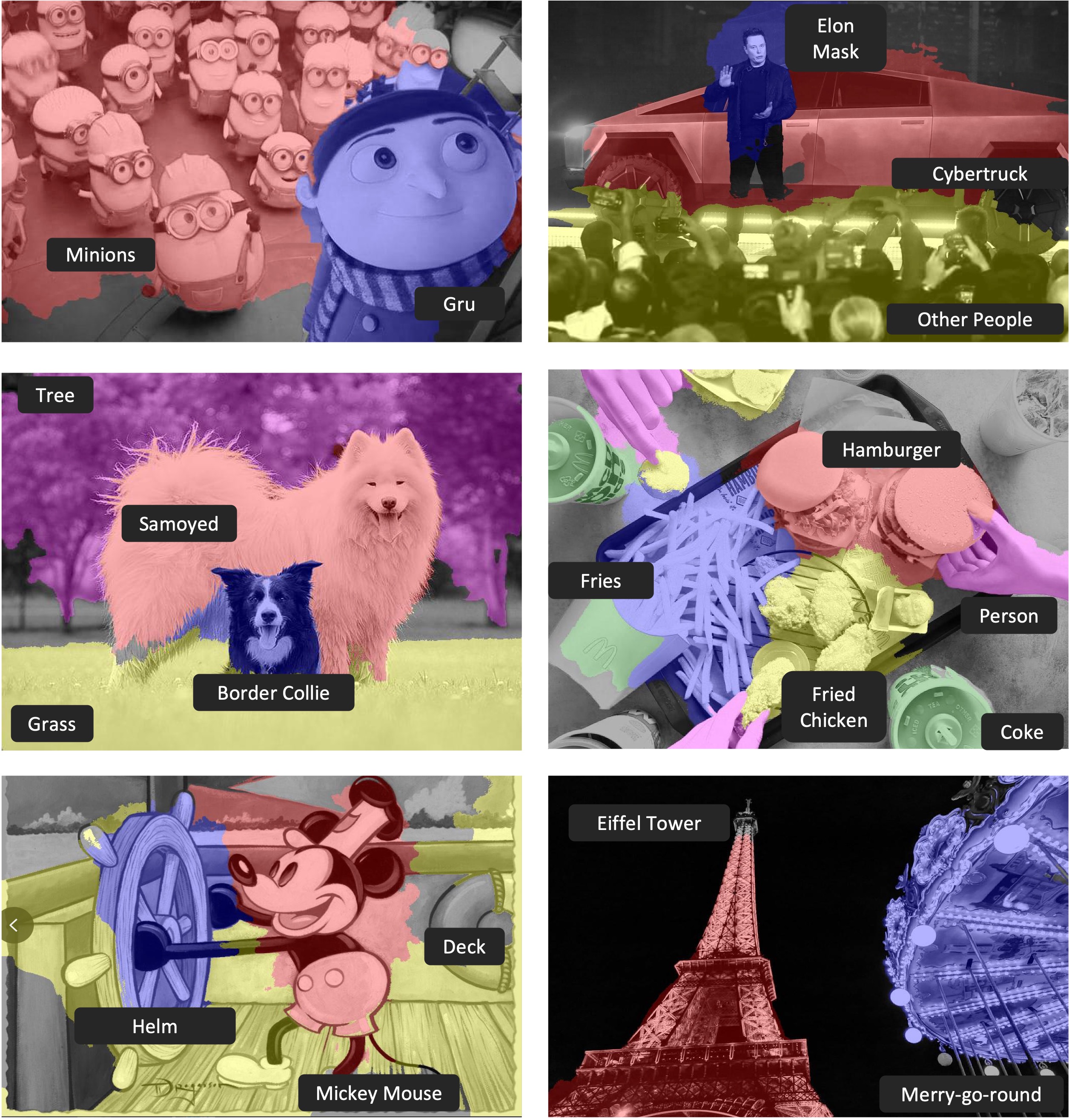}
    
    \caption{\shortname+BLIP$_{\text{Flickr}}$ segmentation result for in the wild images.}
    \label{fig:inthewild}
\end{figure*}

\section{Additional Qualitative Results}

\begin{figure*}
  \centering
    \includegraphics[width=0.95\textwidth]{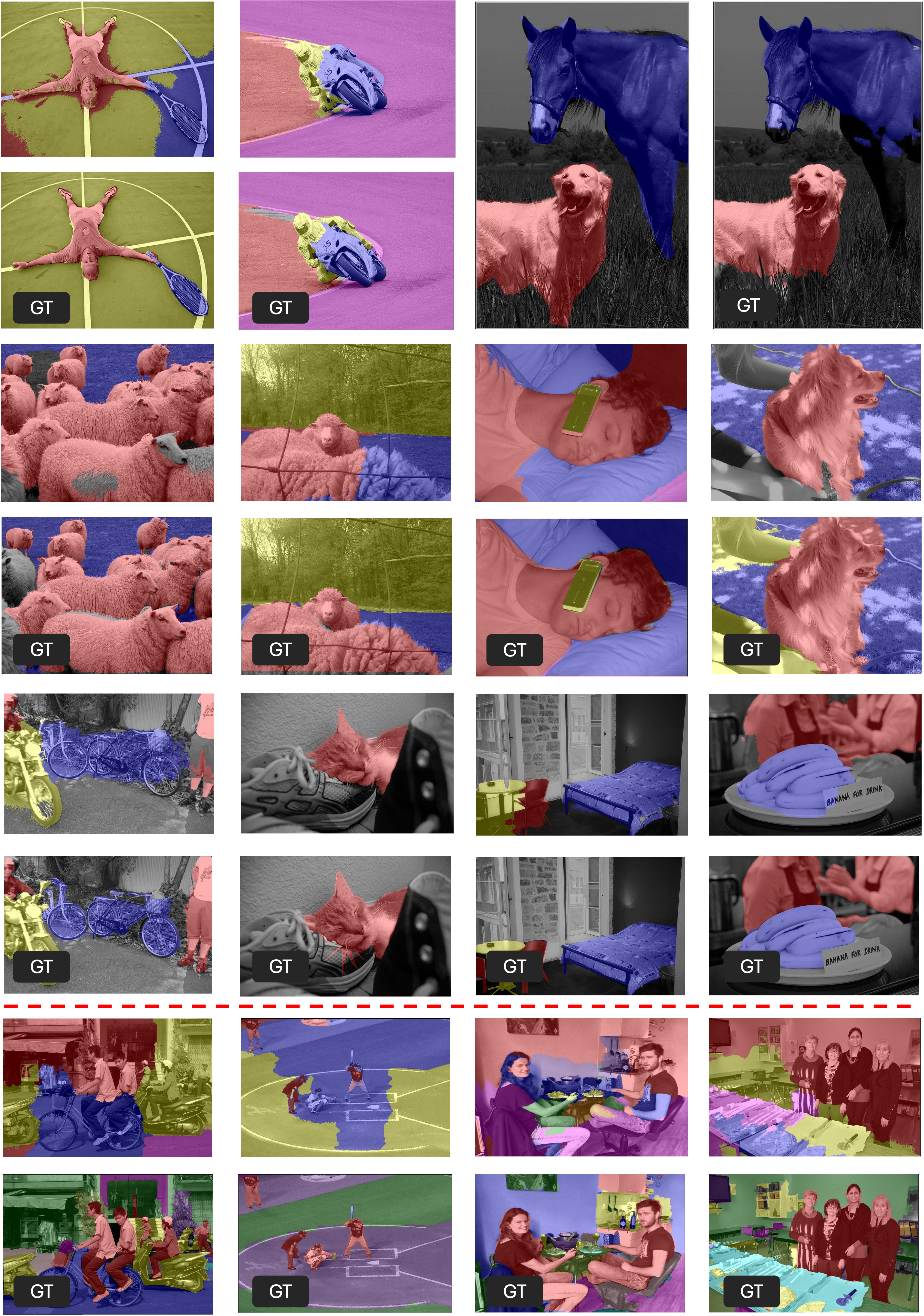}

    \caption{{\bf Qualitative Results of PnP-OVSS + BLIP.} Images are from Pascal Context, COCO Object and COCO stuff. The bottom rows show the ground-truth (GT); the rest are our results. The last row shows failure cases.}
    \label{fig:quali-all}
\end{figure*}

We provide additional qualitative results of images from Pascal Context and COCO Stuff in Fig \ref{fig:quali-all}. 
We note that, in the third image of the third row, \shortname{} correctly recognizes the occluded person missing in the ground truth.

The bottom row are examples of failure cases. 
The examples \st{on top} contains multiple small instances of the same class ({\tt people} and {\tt forks}, which are understandably hard. All the example contains multiple instances of {\tt people}, which causes difficulties for \shortname{} to cover all objects. In addition, images with a clutter of different objects and complex texture often cause a drop in performance.

\section{Qualitative Results in the Wild}

Fig \ref{fig:inthewild}  show qualitative results of \shortname+BLIP$_{\text{Flickr}}$, containing objects not seen in common semantic segmentation datasets, including cartoon characters {\tt Minions} and {\tt Gru},  dog breeds like {\tt Samoyed} and {\tt Border Collie}, food items like {\tt Hamburger}, {\tt Fries}, {\tt Coke}, and {\tt Fried Chicken}, places of interest like {\tt Eiffel Tower} and {\tt Merry-go-round}, a new electronic vehicle {\tt Cybertruck}, and a celebrity {\tt Elon Mask}.

In particular, we would like to point out the difficulty involved in segmenting the Steamboat Willie image, which is in greyscale and has little texture, depriving the network the ability to use color and texture features. Despite that, \shortname{} is able to extract the main components of {\tt Mickey Mouse}, the {\tt Helm} and the {\tt Deck}. 


\section{PnP-OVSS Implementation Detail}
\noindent \textbf{BLIP.} We use the ITM branch of $\text{BLIP}\_{\text{large}}$, which adopts VIT-L/16 as the image encoder, BERT as the text encoder, and insert an extra cross-attention layer for each transformer block of BERT. For each cross-attention layer, the hidden size is 768, and the number of heads is 12. We interpolate the positional embedding to allow input resolution of 768 x 768. We adopt pretrained weights from two checkpoints for image retrieval on COCO and Flickr.

\vspace{0.1in}
\noindent \textbf{BridgeTower.} We use the ITM branch of BridgeTower$\_{large}$, which adopts VIT-L/14 as the image encoder, RoBERTa$\_{large}$ as the text encoder and an 6-layer cross attention encoder. For each cross attention layer of the cross-modal encoder, the hidden size is set to 1,024, and the number of heads is set to 16. We interpolate the positional embedding to allow input resolution of 770 x 770. The model weights are from the {\tt bridgetower\-large-itm-mlm-itc} checkpoint from Huggingface. 

\vspace{0.1in}
\noindent \textbf{Random Search.}
We adopt the random search routine from the Gradient-Free-Optimizers library\footnote{\url{https://github.com/SimonBlanke/Gradient-Free-Optimizers}} \cite{gfo2020} 
with our reward metric (\S \ref{sec:hyperparamtuning}). To parallelize the search process, we divide the search space into three groups and place each group on a GPU card. We perform 34 search iterations in each group. The best hyperparameter set from the three groups with the highest reward is taken as the final search result. 

\vspace{0.1in}
\noindent \textbf{Class Split for Densely Supervised Models.}    
We follow the most common setting \cite{bucher2019zero,gu2020context,pastore2021closer,li2022languagedriven,xu2022simple,zhou2022extract} which save pottedplant, sheep, sofa, train, tvmonitor as the 5 unseen classes for Pascal VOC; cow, motorbike, sofa, cat, boat, fence, bird, tvmonitor, keyboard, aeroplane as 10 unseen classes for Pascal Context; frisbee, skateboard, cardboard, carrot, scissors, suitcase, giraffe, cow, road, wallconcrete, tree, grass, river, clouds, playingfield, as 15 unseen classes for COCO Stuff.

\section{Inference Speed} MaskClip, Reco, and PnP-OVSS take 0.05s, 4.41s, and 2.46s on average, respectively, for inference on a $320\times320$ image. We calculate the inference speeds of the three models on a single A6000 GPU with 48GB of RAM. The results are the average of 20 independent runs. Hence, PnP-OVSS achieves substantially better performance without significant increase in inference time. 


\section{Vil-Seg Evaluation Detail}

In Tab \ref{tab:main result}, different from other methods that require weakly supervised finetuning on image-text data, Vil-Seg is evaluated on subset of datasets. Specifically, the author evaluate their method on 5 classes (potted plant, sheep, sofa, train, tv-monitor) out of the 20 object categories in PASCAL VOC; 4 classes (cow, motorbike, sofa, cat) out of the 59 object categories in PASCAL Context; and 15 classes (frisbee, skateboard, cardboard, carrot, scissors, suitcase, giraffe, cow, road, wall concrete, tree, grass, river, clouds, playing-field) out of the 171 object categories in COCO Stuff dataset.

\section{Details of Zero-shot Semantic Segmentation Techniques}

We summarize current methods for zero-shot semantic segmentation in Tab \ref{tab:related work supp} to enable straightforward comparison between methods. Specifically, we include the supervision used, whether the method require pretraining and finetuning, the pretraining weight, finetuning data and total data size used in each method. 



\section{PnP-OVSS with ALBEF and mPLUG}

As shown in Tab \ref{tab:add result supp}, we further apply PnP-OVSS on two other vision language models with cross-attention and image text matching loss, ALBEF \cite{ALBEF} and mPLUG \cite{li2022mplug}. Nevertheless, the performances are not as good as BLIP or BridgeTower. ALBEF  \cite{ALBEF} is pretrained with only 14M data with ViT-B whereas BLIP and BridgeTower are pretrained with 129M/404M data with ViT-L. We speculate that vision language models require  sufficient numbers of parameters and pretraining data to acquire localization capability. mPLUG \cite{li2022mplug} is another vision language model pretrained with 14M data and ViT-L. However, mPLUG is trained for both image task and video task. With a relatively smaller amount of pretraining data than BLIP or BridgeTower, as well as image and video dual-modality objectives, mPLUG also does not perform well in image object localization.


\begin{table*}
  \centering
  \label{main result}
  \begin{tabular}{@{}llcccr@{}}
    \toprule
    Method & Supervision & Training & PT weight & Finetuning data & PT/T data size \\
    
    \midrule
    \multicolumn{6}{l}{\textbf{\textit{Methods that  require finetuning on dense annotations w/o VL Models}}}\\
    
    SPNet \cite{xian2019semantic} & Pixel+Self & PT+FT & ImageNet & Pascal Voc/COCO stuff & 119K\\
    ZS3Net \cite{bucher2019zero} & Pixel+Self & PT+FT & - & Pascal Voc/ Pascal Context& 5K\\
    CaGNet \cite{gu2020context} & Pixel+Self & PT+FT & - & \makecell{Pascal Voc/Pascal Context\\/COCO stuff} & 123K\\
    STRICT \cite{pastore2021closer} & Pixel+Self & PT+FT & ImageNet & Pascal Voc/COCO stuff & 119K \\
    \midrule
    \multicolumn{6}{l}{\textbf{\textit{Methods that  require finetuning on dense annotations w/ VL Models}}}\\
    
    SimBase \cite{xu2022simple} & Pixel+Self & PT+FT & Maskformer/FCN + CLIP & COCO stuff & 400.1M  \\
    LSeg \cite{li2022languagedriven} & Pixel+Self & PT+FT & ImageNet+CLIP & Pascal VOC/COCO/FSS & 401.3M \\
    MaskCLIP+ \cite{zhou2022extract} & Pixel+Self & PT+FT & CLIP & COCO stuff & 400.1M\\
    \midrule
    \multicolumn{6}{l}{\textbf{\textit{Methods that  require finetuning on image-text pairs}}}\\
    OVSegmentor \cite{xu2023learning}& Text+Self  & PT+FT & DINO+BERT & CC4M & 4.3M \\ 
    Vil-Seg \cite{liu2022open} & Text+Self  & PT+FT & CLIP & CC12M & 412M \\
    GroupVit* \cite{luo2023segclip} & Text & T & - & CC3M+COCO & 3.4M\\
    GroupVit \cite{xu2022groupvit} & Text & T & - & CC12M+YFCC14M & 26M\\
    CLIPpy \cite{ranasinghe2023perceptual}  & Text+Text  & PT+FT & DINO+T5 Sentence & HQITP-134M & 134M    \\
    SegCLIP \cite{luo2023segclip}  & Text+Self  & PT+FT & CLIP & CC3M+COCO & 403.4M \\
    Viewco \cite{ren2023viewco} & Text  & PT+FT & GroupViT & CC12M+YFCC14M  & 26M \\
    TCL\cite{cha2023learning}+PAMR\cite{araslanov2020single} & Text  & PT+FT & CLIP & CC12M+CC3M & 415M \\
    PACL \cite{mukhoti2023open} & Text  & PT+FT & CLIP & \makecell{GCC3M+GCC12M\\+YFCC15M} & 430M \\

    \midrule
    
    \multicolumn{6}{l}{\textbf{\textit{Methods that require finetuning but not image-text pair}}}\\

    MaskCLIP \cite{zhou2022extract} w/ ST & Text+ST & PT+T & CLIP & ImageNet1K & 401.2M\\
    ZeroSeg* \cite{chen2023exploring} & Text+Self & PT+T & CLIP & CC3M+COCO &  403.4M\\
    ZeroSeg \cite{chen2023exploring} & Text+Self & PT+T & CLIP & ImageNet1K &  401.2M\\
    
    \midrule
    \multicolumn{6}{l}{\textbf{\textit{Methods that require no finetuning}}}\\
    MaskCLIP \cite{zhou2022extract} & Text & PT & CLIP & - &  400M\\
    Reco\cite{shin2022reco}& Text & PT & CLIP+ImageNet & - & 400M \\
    PnP-OVSS (Ours)   &   &  \\
    \;\; + BLIP & Text & PT & BLIP$\_{\text{Flickr}}$/BLIP$\_{\text{COCO}}$ & - & 129M \\
    \;\; + BridgeTower & Text & PT & BridgeTower & - & 404M \\
    
    \bottomrule
  \end{tabular}
  \caption{Current methods for zero-shot semantic segmentation. Pixel represents  method require pixel level annotation, Self represents method leverage self supervision, and Text represents method leverage image-text pair annotation. PT stands for Pre-training, T stands for training, ST stands for Self-training, FT stands for finetuning. All methods with pixel supervision are trained on seen categories and tested on unseen categories. For data size, We calculate only the image-caption
    data used for pretraining and all type of data for finetuning.}
  \label{tab:related work supp}
\end{table*}

\begin{table*}[t]
  \centering
  \begin{tabular}{@{}lcccccccc@{}}
    \toprule
    \makecell{Method} & \makecell{Training}  & \makecell{HT on Dense\\Labels } &  \makecell{Short-side \\ Resolution} & \makecell{Pascal\\VOC-20} & \makecell{Pascal\\Context-59} & \makecell{COCO\\Object-80} & \makecell{COCO\\Stuff-171} \\

    \midrule
    PnP-OVSS (Ours)   &   &  \\
    \;\; + ALBEF & $\times$ & $\times$ & 336 & 10.8 & 6.3  &  8.9 & 10.3\\
    \;\; + ALBEF & $\times$ & $\times$ & 768 & 11.1 & 6.8  & 8.8  & 10.7\\
    \;\; + mPLUG & $\times$ & $\times$ & 336 & 9.2 & 8.8  & 7.9  & 6.7\\

   \bottomrule
  \end{tabular}
  \caption{Zero-shot semantic segmentation performance in mIoU. }
  \label{tab:add result supp}
\end{table*}

\end{document}